\documentclass{article}

\usepackage{etoolbox}
\newtoggle{arxiv}
\toggletrue{arxiv}

\newtoggle{icmlworkshop}
\togglefalse{icmlworkshop}

\iftoggle{arxiv}{}{
\PassOptionsToPackage{numbers,sort,compress}{natbib}
\usepackage{neurips_2022}
}

\usepackage[utf8]{inputenc} %
\usepackage[T1]{fontenc}    %
\usepackage{hyperref}       %
\usepackage{url}            %
\usepackage{booktabs}       %
\usepackage{amsfonts}       %
\usepackage{nicefrac}       %
\usepackage{microtype}      %
\usepackage{xcolor}         %
\usepackage{amsmath,amsthm,amssymb}

\usepackage{algorithm}
\usepackage{algorithmic}

\usepackage{subcaption}
\usepackage{multirow}

\usepackage{xspace}
\usepackage{wrapfig}

\usepackage{pifont}
\newcommand{\xmark}{\ding{55}}

\usepackage[capitalise]{cleveref}  %
\usepackage{comment}
\usepackage[inline]{enumitem}

\usepackage{graphicx}
\usepackage{grffile}
\usepackage{color}

\usepackage{newtxmath}

\iftoggle{arxiv}{
\usepackage[numbers,sort]{natbib}
}{}

\usepackage{import}

\newtoggle{todo}
\toggletrue{todo}

\newtoggle{comment}
\togglefalse{comment} %

\newcommand{\diag}{\mathrm{diag}}
\newcommand{\softmax}{\mathrm{softmax}}

\newcommand{\defeq}{:=}

\newcommand{\vQ}{\mathbf{Q}}
\newcommand{\vK}{\mathbf{K}}
\newcommand{\vV}{\mathbf{V}}
\newcommand{\vdQ}{\mathbf{dQ}}
\newcommand{\vdK}{\mathbf{dK}}
\newcommand{\vdV}{\mathbf{dV}}
\newcommand{\vS}{\mathbf{S}}
\newcommand{\vdS}{\mathbf{dS}}
\newcommand{\vP}{\mathbf{P}}
\newcommand{\vdP}{\mathbf{dP}}

\newcommand{\vO}{\mathbf{O}}
\newcommand{\vdO}{\mathbf{dO}}
\newcommand{\vM}{\mathbf{M}}
\newcommand{\vZ}{\mathbf{Z}}

\newcommand{\sysname}{\textsc{FlashAttention}\xspace}

\newtheorem{theorem}{Theorem}
\newtheorem*{theorem*}{Theorem}

\newtheorem{proposition}[theorem]{Proposition}

\iftoggle{arxiv}{
  \setlength{\textwidth}{6.5in}
  \setlength{\textheight}{9in}
  \setlength{\oddsidemargin}{0in}
  \setlength{\evensidemargin}{0in}
  \setlength{\topmargin}{-0.5in}
  \newlength{\defbaselineskip}
  \setlength{\defbaselineskip}{\baselineskip}
  \setlength{\marginparwidth}{0.8in}
}{
\usepackage[compact]{titlesec}
\titlespacing{\section}{0pt}{*1}{*0}
\titlespacing{\subsection}{0pt}{*1.5}{*0}

\usepackage[subtle, mathdisplays=tight, charwidths=normal, leading=normal]{savetrees}

\addtolength\textfloatsep{-0.5em}
\addtolength\intextsep{-0.2em}

\def\setstretch#1{\renewcommand{\baselinestretch}{#1}}
\setstretch{0.985}
\addtolength{\parskip}{-1pt}

}

\title{\sysname: Fast and Memory-Efficient Exact Attention with IO-Awareness}

\iftoggle{arxiv}{
  \usepackage{authblk}
  \author[$\dagger$]{Tri Dao}
  \author[$\dagger$]{Daniel Y.\ Fu}
  \author[$\dagger$]{Stefano Ermon}
  \author[$\ddagger$]{Atri Rudra}
  \author[$\dagger$]{Christopher R{\'e}}
  \affil[$\dagger$]{Department of Computer Science, Stanford University}
  \affil[$\ddagger$]{Department of Computer Science and Engineering, University at Buffalo, SUNY\vspace{4pt}}
  \affil[ ]{{\texttt{\{trid,danfu\}@cs.stanford.edu}, \texttt{ermon@stanford.edu}, \texttt{atri@buffalo.edu}, \texttt{chrismre@cs.stanford.edu}}}
}{

\author{%
  Tri Dao$^{\dagger}$, Daniel Y.\ Fu $^\dagger$, Stefano Ermon $^\dagger$, Atri Rudra $^\ddagger$, Christopher R\'{e} $^\dagger$\\
  $^\dagger$ Department of Computer Science, Stanford University\\
  $^\ddagger$ Department of Computer Science and Engineering, University at Buffalo, SUNY\\
  {\small\texttt{\{trid,danfu\}@stanford.edu}, \texttt{ermon@stanford.edu}, \texttt{atri@buffalo.edu}, \texttt{chrismre@cs.stanford.edu}}
}
}

\begin{document}

\maketitle

\begin{abstract}

Transformers are slow and memory-hungry on long sequences, since the time and memory complexity of self-attention are quadratic in sequence length.
Approximate attention methods have attempted to address this problem by
trading off model quality to reduce the compute complexity, but often do not achieve wall-clock speedup.
We argue that a missing principle is making attention algorithms \textit{IO-aware}---accounting for reads and writes between levels of GPU memory.
We propose \sysname, an IO-aware exact attention algorithm that uses tiling to reduce the number of memory reads/writes between GPU high bandwidth memory (HBM) and GPU on-chip SRAM.
We analyze the IO complexity of \sysname, showing that it requires fewer HBM accesses than standard attention, and is optimal for a range of SRAM sizes.
We also extend \sysname to block-sparse attention, yielding an approximate attention algorithm that is faster than any existing approximate attention method.
\sysname trains Transformers faster than existing baselines: 15\% end-to-end wall-clock speedup on BERT-large (seq.\ length 512) compared to the MLPerf 1.1 training speed record, 3$\times$ speedup on GPT-2 (seq.\ length 1K), and 2.4$\times$ speedup on long-range arena (seq.\ length 1K-4K).
\sysname and block-sparse \sysname enable longer context in Transformers, yielding higher quality models (0.7 better perplexity on GPT-2 and 6.4 points of lift on long-document classification) and entirely new capabilities: the first Transformers to achieve better-than-chance performance on the Path-X challenge (seq.\ length 16K, 61.4\% accuracy) and Path-256 (seq.\ length 64K, 63.1\% accuracy).

\end{abstract}

\section{Introduction}
\label{sec:intro}

Transformer models~\citep{vaswani2017attention} have emerged as the most widely used architecture in applications such as natural language processing and image classification.
Transformers have grown larger~\citep{brown2020language} and deeper~\citep{wang2022deepnet}, but equipping them with longer context remains difficult~\citep{tay2020long}, since the self-attention module at their heart has time and memory complexity quadratic in sequence length.
An important question is whether making attention faster and more memory-efficient can help Transformer models address their runtime and memory challenges for long sequences.

Many approximate attention methods have aimed to reduce the compute and memory requirements of attention.
These methods range from sparse-approximation~\citep{kitaev2020reformer, roy2021efficient} to low-rank approximation~\citep{wang2020linformer, katharopoulos2020transformers, choromanski2020rethinking}, and their combinations~\citep{beltagy2020longformer, zaheer2020bigbird, scatterbrain}.
Although these methods reduce the compute requirements to linear or near-linear in sequence length, many of them do not display wall-clock speedup against standard attention and have not gained wide adoption. One main reason is that they focus on FLOP reduction (which may not correlate with wall-clock speed) and tend to ignore overheads from memory access (IO).

\iftoggle{icmlworkshop}{
\begin{figure*}[t]
}{
\begin{figure}[t]
}
\centering
\includegraphics[width=5.5in]{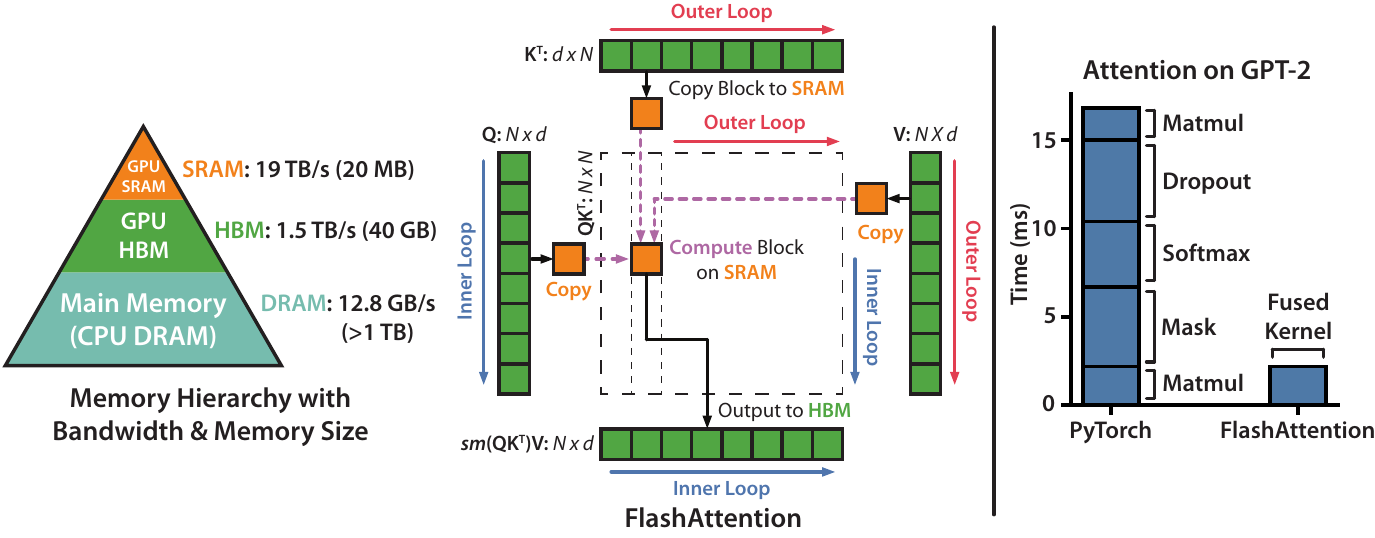}
\caption{
\textbf{Left:} \sysname uses tiling to prevent materialization of the large $N \times N$ attention matrix (dotted box) on (relatively) slow GPU HBM. In the outer loop (red arrows), \sysname loops through blocks of the $\vK$ and $\vV$ matrices and loads them to fast on-chip SRAM.
In each block, \sysname loops over blocks of $\vQ$ matrix (blue arrows), loading them to SRAM, and writing the output of the attention computation back to HBM.
\textbf{Right:} Speedup over the PyTorch implementation of attention on GPT-2.
\sysname does not read and write the large $N\times N$ attention matrix to HBM, resulting in an 7.6$\times$ speedup on the attention computation.}
\label{fig:banner}
\iftoggle{arxiv}{}{
\vspace{-2em}
}
\iftoggle{icmlworkshop}{
\end{figure*}
}{
\end{figure}
}

In this paper, we argue that a missing principle is making attention algorithms \textit{IO-aware}~\citep{aggarwal1988input}---that is, carefully accounting for reads and writes to different levels of fast and slow memory (e.g., between fast GPU on-chip SRAM and relatively slow GPU high bandwidth memory, or HBM~\citep{jia2018dissecting}, Figure~\ref{fig:banner} left).
On modern GPUs, compute speed has out-paced memory speed~\citep{nvidia2017nvidia,nvidia2020nvidia,nvidia2022nvidia}, and most operations in Transformers are bottlenecked by memory accesses~\citep{ivanov2021data}.
IO-aware algorithms have been critical for similar memory-bound operations, when reading and writing data can account for a large portion of the runtime---such as database joins~\citep{ramakrishnan2003database}, image processing~\citep{ragan2013halide}, numerical linear algebra~\citep{blackford2002updated}, and more~\citep{williams2009roofline, hennessy2003memory}.
However, common Python interfaces to deep learning such as PyTorch and Tensorflow do not allow fine-grained control of memory access.

We propose \sysname, a new attention algorithm that computes exact attention with far fewer memory accesses.
Our main goal is to avoid reading and writing the attention matrix to and from HBM.
This requires (i) computing the softmax reduction without access to the whole input (ii) not storing the large intermediate attention matrix for the backward pass.
We apply two well-established techniques to address these challenges.
(i) We restructure the attention computation to split the input into blocks and make several passes over input blocks, thus incrementally performing the softmax reduction (also known as \textbf{tiling}). (ii) We store the softmax normalization factor from the forward pass to quickly \textbf{recompute} attention on-chip in the backward pass, which is faster than the standard approach of reading the intermediate attention matrix from HBM.
We implement \sysname in CUDA to achieve fine-grained control over memory access and fuse all the attention operations into one GPU kernel.
Even with the increased FLOPs due to recomputation, our algorithm both \textbf{runs faster} (up to 7.6x on GPT-2~\citep{radford2019language}, Figure~\ref{fig:banner} right) and \textbf{uses less memory}---linear in sequence length---than standard attention, thanks to the massively reduced amount of HBM access.

We analyze the IO complexity~\citep{aggarwal1988input} of \sysname, proving that it requires $O(N^2 d^2 M^{-1})$ HBM accesses where $d$ is the head dimension and $M$ is the size of SRAM, as compared to $\Omega(Nd + N^2)$ of standard attention.
For typical values of $d$ and $M$, \sysname requires many times fewer HBM accesses compared to standard attention (up to 9$\times$ fewer, as shown in~\cref{fig:micros}).
Moreover, we provide a lower bound, showing that no exact attention algorithm can asymptotically improve on the number of HBM accesses over all SRAM sizes.

We also show that \sysname can serve as a 
useful primitive for realizing the potential of approximate attention algorithms by overcoming their issues with memory access overhead.
As a proof of concept, we implement block-sparse \sysname, a sparse attention algorithm that is 2-4$\times$ faster than even \sysname, scaling up to sequence length of 64k.
We prove that block-sparse \sysname has better IO complexity than \sysname by a factor proportional to the sparsity ratio.
We discuss further extensions to other operations (attention on multi-GPU, kernel regression, block-sparse matrix multiply) in~\cref{sec:discussion}.
We open-source \sysname to make it easier to build on this primitive.\footnote{\sysname code is available at \url{https://github.com/HazyResearch/flash-attention}}

We empirically validate that \sysname speeds up model training and improves model quality by modeling longer context. We also benchmark the runtime and memory footprint of \sysname and block-sparse \sysname compared to prior attention implementations.
\begin{itemize}[itemsep=0.1pt,topsep=0pt,leftmargin=*]
  \item \textbf{Faster Model Training.} \sysname trains Transformer models faster in wall-clock time. We train BERT-large (seq.\ length 512) 15\% faster than the training speed record in MLPerf 1.1~\citep{mattson2020mlperf}, GPT2 (seq.\ length 1K) 3$\times$ faster than baseline implementations from HuggingFace~\citep{wolf-etal-2020-transformers} and Megatron-LM~\citep{shoeybi2019megatron}, and long-range arena (seq.\ length 1K-4K) 2.4$\times$ faster than baselines.
  \item \textbf{Higher Quality Models.} \sysname scales Transformers to longer sequences, which improves their quality and enables new capabilities.
  We observe a 0.7 improvement in perplexity on GPT-2 and 6.4 points of lift from modeling longer sequences on long-document classification~\citep{dai2022revisiting}.
  \sysname enables the first Transformer that can achieve better-than-chance performance on the Path-X~\citep{tay2020long} challenge, solely from using a longer sequence length (16K).
  Block-sparse \sysname enables a Transformer to scale to even longer sequences (64K), resulting in the first model that can achieve better-than-chance performance on Path-256.
  \item \textbf{Benchmarking Attention.} \sysname is up to 3$\times$ faster than the standard attention implementation across common sequence lengths from 128 to 2K and scales up to 64K.
  Up to sequence length of 512, \sysname is both faster and more memory-efficient than any existing attention method, whereas for sequence length beyond 1K, some approximate attention methods (e.g., Linformer) start to become faster.
  On the other hand, block-sparse \sysname is faster than all existing approximate attention methods that we know of.
  
\end{itemize}

\section{Background}
\label{sec:background}

We provide some background on the performance characteristics of common deep
learning operations on modern hardware (GPUs).
We also describe the standard implementation of attention.

\subsection{Hardware Performance}
\label{subsec:hardware}

We focus here on GPUs.
Performance on other hardware accelerators are similar~\citep{jouppi2017datacenter, jia2019dissecting}.

\textbf{GPU Memory Hierarchy.}
The GPU memory hierarchy (\cref{fig:banner} left) comprises multiple forms of memory of different
sizes and speeds, with smaller memory being faster.
As an example, the A100 GPU has 40-80GB of high bandwidth memory (HBM) with
bandwidth 1.5-2.0TB/s and 192KB of on-chip SRAM per each of 108 streaming
multiprocessors with
bandwidth estimated around 19TB/s~\citep{jia2018dissecting, jia2021dissecting}.
The on-chip SRAM is an order of magnitude faster than HBM but many orders of
magnitude smaller in size.
As compute has gotten faster relative to memory speed~\citep{nvidia2017nvidia,nvidia2020nvidia,nvidia2022nvidia}, operations
are increasingly bottlenecked by memory (HBM) accesses.
Thus exploiting fast SRAM becomes more important.

\textbf{Execution Model.}
GPUs have a massive number of threads to execute an operation
(called a kernel).
Each kernel loads inputs from HBM to registers and SRAM, computes, then writes outputs to HBM.

\textbf{Performance characteristics.} Depending on the balance of computation and memory accesses, operations can be
classified as either compute-bound or memory-bound.
This is commonly measured by the \emph{arithmetic intensity}~\citep{williams2009roofline},
which is the number of arithmetic operations per byte of memory access.
\begin{enumerate}[itemsep=0.1pt,topsep=0pt,leftmargin=*]
  \item Compute-bound: the time taken by the operation is determined by how many
  arithmetic operations there are, while time accessing HBM
  is much smaller. Typical examples are matrix multiply with large inner
  dimension, and convolution with large number of channels.
  \item Memory-bound: the time taken by the operation is determined by the
  number of memory accesses, while time spent in computation is much smaller.
  Examples include most other operations:
  elementwise (e.g., activation, dropout), and reduction (e.g., sum,
  softmax, batch norm, layer norm).
\end{enumerate}

\textbf{Kernel fusion.}
The most common approach to accelerate memory-bound operations is
kernel fusion: if there are multiple operations applied to the same input,
the input can be loaded once from HBM, instead of multiple times for each operation.
Compilers can automatically fuse many elementwise operations~\citep{li2020deep, paszke2019pytorch, sabne2020xla}.
However, in the context of model training, the intermediate values still need
to be written to HBM to save for the backward pass, reducing the
effectiveness of naive kernel fusion.

\subsection{Standard Attention Implementation}
\label{subsec:standard_attn}

Given input sequences $\vQ, \vK, \vV \in \mathbb{R}^{N \times d}$ where $N$ is the sequence length and
$d$ is the head dimension, we want to compute the attention output $\vO \in \mathbb{R}^{N \times d}$:
\begin{equation*}
  \vS = \vQ \vK^\top \in \mathbb{R}^{N \times N}, \quad \vP = \softmax(\vS) \in \mathbb{R}^{N \times N}, \quad \vO = \vP\vV \in \mathbb{R}^{N \times d},
\end{equation*}
where $\softmax$ is applied row-wise.

Standard attention implementations materialize the matrices $\vS$ and $\vP$ to HBM, which takes $O(N^2)$ memory.
Often $N \gg d$ (e.g., for GPT2, $N = 1024$ and $d = 64$).
We describe the standard attention implementation in~\cref{alg:standard_attn}.
As some or most of the operations are memory-bound (e.g., softmax), the large number of
memory accesses translates to slow wall-clock time.

This problem is exacerbated by other elementwise operations applied
to the attention matrix, such as masking applied to $\vS$ or dropout applied to $\vP$.
As a result, there have been many attempts to fuse several elementwise
operations, such as fusing masking with softmax~\citep{shoeybi2019megatron}.

In \cref{sec:theory}, we will show that the standard attention implementation
performs HBM accesses quadratic in the sequence length $N$.
We also compare the number of FLOPs and number of HBM accesses of standard
attention and of our method (\sysname).

\vspace{-0.5em}
\setcounter{algorithm}{-1}
\begin{algorithm}[H]
  \caption{\small\label{alg:standard_attn}Standard Attention Implementation}
  \begin{algorithmic}[1]
    \REQUIRE Matrices $\vQ, \vK, \vV \in \mathbb{R}^{N \times d}$ in HBM.
    \STATE \label{alg:standard_attn_qk} Load $\vQ, \vK$ by blocks from HBM, compute
    $\vS = \vQ \vK^\top$, write $\vS$ to HBM.
    \STATE \label{alg:standard_attn_sp} Read $\vS$ from HBM, compute $\vP = \softmax(\vS)$, write $\vP$ to
    HBM.
    \STATE \label{alg:standard_attn_pv} Load $\vP$ and $\vV$ by blocks from HBM, compute $\vO = \vP\vV$, write $\vO$ to
    HBM.
    \STATE Return $\vO$.
  \end{algorithmic}
\end{algorithm}
\vspace{-1.0em}

\iftoggle{icmlworkshop}{
\section{\sysname: Efficient Attention with Tiling \& Recomputation}
}{
\section{\sysname: Algorithm, Analysis, and Extensions}
}
\label{sec:algo}

We show how to compute exact attention with fewer HBM reads/writes and without storing large intermediate matrices for the backward pass.
This yields an attention algorithm that is both memory efficient and faster
in wall-clock time.
We analyze its IO complexity, showing that our method requires
much fewer HBM accesses compared to standard attention.
We further show that \sysname can serve as a useful primitive by extending it to handle block-sparse attention.

\iftoggle{icmlworkshop}{
\cref{sec:background} contains background on GPU architecture and the performance characteristics of common deep learning operations.
We note that attention is memory-bound: its runtime is bottlenecked by the time taken by HBM reads/writes.
}{}
We focus here on the forward pass for ease of exposition; \cref{sec:algo_details} contains details
for the backward.

\iftoggle{icmlworkshop}{}{
\subsection{An Efficient Attention Algorithm With Tiling and Recomputation}
\label{sec:implementation}
}

Given the inputs $\vQ, \vK, \vV \in \mathbb{R}^{N \times d}$ in HBM, we aim to compute the attention output $\vO \in \mathbb{R}^{N \times d}$ and write it to HBM.
Our goal is to reduce the amount of HBM accesses (to sub-quadratic in $N$).

We apply two established techniques (tiling, recomputation) to overcome the
technical challenge of computing exact attention in sub-quadratic HBM accesses.
We describe this in~\cref{alg:stream_attn}.
The main idea is that we split the inputs $\vQ, \vK, \vV$ into blocks,
load them from slow HBM to fast SRAM, then compute the attention output with
respect to those blocks.
By scaling the output of each block by the right normalization factor before
adding them up, we get the correct result at the end.

\textbf{Tiling.}
We compute attention by blocks.
Softmax couples columns of $\vK$,
so we decompose the large softmax
with scaling~\citep{milakov2018online, kitaev2020reformer, rabe2021self}.
\iftoggle{icmlworkshop}{}{
For numerical stability, the softmax of vector $x \in \mathbb{R}^{B}$ is computed as:
\begin{equation*}
  m(x) \defeq \max_i\ \ x_i, \quad
  f(x) \defeq \begin{bmatrix} e^{x_1 - m(x)} & \hdots & e^{x_B - m(x)} \end{bmatrix}, \quad
  \ell(x) \defeq \sum_i f(x)_i, \quad
  \softmax(x) \defeq \frac{f(x)}{\ell(x)}.
\end{equation*}
For vectors $x^{(1)}, x^{(2)} \in \mathbb{R}^{B}$, we can decompose the softmax of the concatenated $x = \begin{bmatrix} x^{(1)} \ x^{(2)} \end{bmatrix} \in \mathbb{R}^{2B}$ as:
\begin{align*}
  &m(x) = m(\begin{bmatrix} x^{(1)} \ x^{(2)} \end{bmatrix}) = \max(m(x^{(1)}), m(x^{(2)})), \quad
  f(x) = \begin{bmatrix} e^{m(x^{(1)}) - m(x)} f(x^{(1)}) & e^{m(x^{(2)}) - m(x)} f(x^{(2)}) \end{bmatrix}, \\
  &\ell(x) = \ell(\begin{bmatrix} x^{(1)} \ x^{(2)} \end{bmatrix}) = e^{m(x^{(1)}) - m(x)}\ell (x^{(1)}) + e^{m(x^{(2)}) - m(x)} \ell(x^{(2)}), \quad
  \softmax(x) = \frac{f(x)}{\ell(x)}.
\end{align*}
Therefore if we keep track of some extra statistics ($m(x), \ell(x)$), we can compute softmax one block at a time.\footnote{This style of aggregation is called \emph{algebraic aggregation}~\citep{gray1997data}.}
}
We thus split the inputs $\vQ, \vK, \vV$ into blocks (\cref{alg:stream_attn} line \ref{alg:stream_attn_split_qkv}), compute the softmax values along with extra statistics (\cref{alg:stream_attn} line \ref{alg:stream_attn_statistics}), and combine the results (\cref{alg:stream_attn} line \ref{alg:stream_attn_aggregate}).

\textbf{Recomputation.}
One of our goals is to not store $O(N^2)$ intermediate values for the backward
pass.
The backward pass typically requires the matrices
$\vS, \vP \in \mathbb{R}^{N \times N}$ to compute the gradients with respect to $\vQ, \vK, \vV$.
However, by storing the output $\vO$ and the softmax normalization statistics $(m, \ell)$, we can
recompute the attention matrix $\vS$ and $\vP$ easily in the backward pass from blocks of $\vQ, \vK, \vV$ in SRAM.
This can be seen as a form of selective gradient checkpointing~\citep{griewank2008evaluating, chen2016training}.
While gradient checkpointing has been suggested to reduce the maximum amount of memory required~\citep{rabe2021self}, all implementations (that we know off) have to trade speed for memory.
In contrast, even with more FLOPs, our recomputation speeds up the backward pass due to reduced HBM accesses (\cref{fig:micros}).
The full backward pass description is in~\cref{sec:algo_details}.

\textbf{Implementation details: Kernel fusion.}
Tiling enables us to implement our algorithm in one CUDA kernel, loading input from HBM,
performing all the computation steps (matrix multiply, softmax, optionally
masking and dropout, matrix multiply), then write the result back to HBM (masking and dropout in~\cref{sec:algo_details}).
This avoids repeatedly reading and writing of inputs and outputs from and to HBM.

\vspace{-0.5em}
\begin{algorithm}[H]
  \iftoggle{icmlworkshop}{
  \small
  }{}
  \caption{\small\label{alg:stream_attn}\sysname}
  \begin{algorithmic}[1]
    \REQUIRE Matrices $\vQ, \vK, \vV \in \mathbb{R}^{N \times d}$ in HBM, on-chip SRAM of
    size $M$.
    \STATE Set block sizes $B_c = \left\lceil \frac{M}{4d} \right\rceil, B_r = \min \left( \left\lceil \frac{M}{4d} \right\rceil , d \right)$.
    \STATE \label{alg:stream_attn_init} Initialize $\vO = (0)_{N \times d} \in \mathbb{R}^{N \times d}, \ell = (0)_N \in \mathbb{R}^{N}, m = (-\infty)_N \in \mathbb{R}^{N}$ in HBM.
    \STATE \label{alg:stream_attn_split_qkv} Divide $\vQ$ into $T_r = \left\lceil\frac{N}{B_r} \right\rceil$ blocks $\vQ_1, \dots, \vQ_{T_r}$ of size $B_r \times d$ each,
    and divide $\vK, \vV$ in to $T_c = \left\lceil \frac{N}{B_c} \right\rceil$ blocks $\vK_1, \dots, \vK_{T_c}$ and
    $\vV_1, \dots, \vV_{T_c}$, of size $B_c \times d$ each.
    \STATE Divide $\vO$ into $T_r$ blocks $\vO_i, \dots, \vO_{T_r}$ of size
    $B_r \times d$ each, divide $\ell$ into $T_r$ blocks $\ell_i, \dots, \ell_{T_r}$ of size
    $B_r$ each, divide $m$ into $T_r$ blocks $m_1, \dots, m_{T_r}$ of size $B_r$ each.
    \FOR{$1 \le j \le T_c$} \label{alg:stream_attn_outer_loop}
      \STATE \label{alg:stream_attn_load_kv} Load $\vK_j, \vV_j$ from HBM to on-chip SRAM.
      \FOR{$1 \le i \le T_r$}
        \STATE \label{alg:stream_attn_load_qo} Load $\vQ_i, \vO_i, \ell_i, m_i$ from HBM to on-chip SRAM.
        \STATE \label{alg:stream_attn_qk} On chip, compute $\vS_{ij} = \vQ_i \vK_j^T \in \mathbb{R}^{B_r \times B_c}$.
        \STATE \label{alg:stream_attn_statistics} On chip, compute $\tilde{m}_{ij} = \mathrm{rowmax}(\vS_{ij}) \in \mathbb{R}^{B_r}$, $\tilde{\vP}_{ij} = \exp(\vS_{ij} - \tilde{m}_{ij}) \in \mathbb{R}^{B_r \times B_c}$ (pointwise),
        $\tilde{\ell}_{ij} = \mathrm{row sum}(\tilde{\vP}_{ij}) \in \mathbb{R}^{B_r}$.
        \STATE On chip, compute $m_i^{\mathrm{new}} = \max(m_i, \tilde{m}_{ij}) \in \mathbb{R}^{B_r}$, $\ell_i^{\mathrm{new}} = e^{m_i - m_i^{\mathrm{new}}} \ell_i + e^{\tilde{m}_{ij} - m_i^{\mathrm{new}}} \tilde{\ell}_{ij} \in \mathbb{R}^{B_r}$.
        \STATE \label{alg:stream_attn_aggregate} Write $\vO_i \leftarrow \diag(\ell_i^{\mathrm{new}})^{-1}(\diag(\ell_i) e^{m_i - m_i^{\mathrm{new}}} \vO_i + e^{\tilde{m}_{ij} - m_i^{\mathrm{new}}}\tilde{\vP}_{ij} \vV_j)$
        to HBM.
        \STATE Write $\ell_i \leftarrow \ell_i^{\mathrm{new}}$, $m_i \leftarrow m_i^{\mathrm{new}}$ to HBM.
      \ENDFOR
    \ENDFOR
    \STATE Return $\vO$.
  \end{algorithmic}
\end{algorithm}
\vspace{-0.5em}

We show \sysname's correctness, runtime, and memory requirement (proof in~\cref{sec:proofs}).
\begin{theorem}
  \label{thm:correctness}
  \cref{alg:stream_attn} returns $\vO = \softmax(\vQ\vK^\top)\vV$ with $O(N^2d)$ FLOPs and
  requires $O(N)$ additional memory beyond inputs and output.
\end{theorem}

\iftoggle{icmlworkshop}{
In~\cref{sec:theory}, we analyze the IO-complexity of \sysname, proving that it
requires fewer HBM accesses than standard attention, and the complexity is
optimal for a range of SRAM size $M$.
We discuss various extensions to \sysname in~\cref{sec:blocks_sparse,sec:extension_details}.
}{}

\iftoggle{icmlworkshop}{
\section{Analysis: IO Complexity of \sysname}
}
\subsection{Analysis: IO Complexity of \sysname}
\label{sec:theory}

We analyze the IO complexity of \sysname, showing
significant reduction in HBM accesses compared to standard attention.
We also provide a lower bound, proving that no exact attention algorithm can asymptotically improve on HBM accesses over all
SRAM sizes.
Proofs are in \cref{sec:proofs}.

\begin{theorem}\label{thm:io_complexity}
  Let $N$ be the sequence length, $d$ be the head dimension, and $M$ be size of
  SRAM with $d \leq M \leq Nd$.
  Standard attention (\cref{alg:standard_attn}) requires $\Theta(Nd + N^2)$ HBM
  accesses, while \sysname (\cref{alg:stream_attn}) requires
  $\Theta ( N^2 d^2 M^{-1} )$ HBM accesses.
\end{theorem}
For typical values of $d$ (64-128) and $M$ (around 100KB), $d^2$ is many
times smaller than $M$, and thus \sysname requires many times fewer
HBM accesses than standard implementation.
This leads to both faster execution and lower memory footprint, which we
validate in~\cref{sec:benchmark}.

The main idea of the proof is that given the SRAM size of $M$, we
can load blocks of $\vK, \vV$ of size $\Theta(M)$ each (\cref{alg:stream_attn} line \ref{alg:stream_attn_load_kv}).
For each block of $\vK$ and $\vV$, we iterate over all blocks of $\vQ$
(\cref{alg:stream_attn} line \ref{alg:stream_attn_load_qo}) to compute the
intermediate values, resulting in $\Theta(NdM^{-1})$ passes over $\vQ$.
Each pass loads $\Theta(Nd)$ elements, which amounts to $\Theta(N^2 d^2 M^{-1})$ HBM accesses.
We similarly prove that the backward pass of standard attention requires
$\Theta(Nd + N^2)$ HBM accesses while the backward pass of \sysname requires
$\Theta(N^2 d^2 M^{-1})$ HBM accesses (\cref{sec:algo_details}).

We prove a lower-bound: one cannot asymptotically improve on the number of HBM
accesses for all values of $M$ (the SRAM size) when computing exact attention.
\begin{proposition}\label{thm:lower_bound}
  Let $N$ be the sequence length, $d$ be the head dimension, and $M$ be size of
  SRAM with $d \leq M \leq Nd$.
  There does not exist an algorithm to compute exact attention with
  $o(N^2d^2 M^{-1})$ HBM accesses for all $M$ in the range
  $[d, Nd]$.
\end{proposition}
The proof relies on the fact that for $M = \Theta(Nd)$ any algorithm must perform
$\Omega ( N^2d^2M^{-1} ) = \Omega(Nd)$ HBM accesses.
This type of lower bound over a subrange of $M$ is common in the streaming
algorithms literature~\citep{woodruff2004optimal}.
We leave proving parameterized complexity~\citep{flum2006parameterized} lower
bounds in terms of $M$ as exciting future work.

\begin{figure}[t]
  \captionsetup{font=small}
    \centering
    \begin{minipage}{2.3in}
        \centering
        \resizebox{0.98\linewidth}{!}
        {
        \begin{tabular}{@{}c|ccc@{}}
          Attention & Standard & \sysname \\ \hline
          GFLOPs & 66.6 & 75.2 \\
          HBM R/W (GB) & 40.3 & 4.4 \\
          Runtime (ms) & 41.7 & 7.3
        \end{tabular}
        }
    \end{minipage}
    \begin{minipage}{3in}
        \centering
        \includegraphics[width=3in]{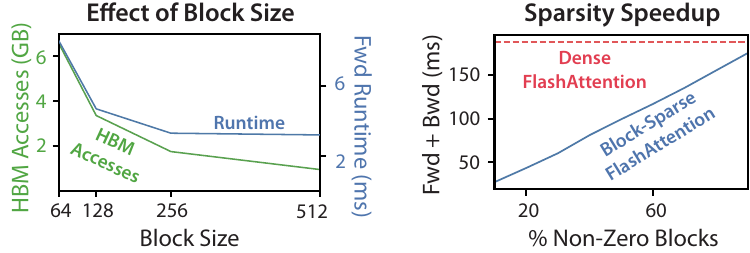}
    \end{minipage}
    \captionsetup{font=small}
    \caption{\label{fig:micros}
    \textbf{Left}: Forward + backward runtime of
    standard attention and \sysname for GPT-2 medium
    (seq.\ length 1024, head dim.\ 64, 16 heads, batch size 64) on
    A100 GPU.
    HBM access is the primary factor affecting runtime.
    \textbf{Middle}: Forward runtime of \sysname
    (seq.\ length 1024, head
    dim.\ 64, 16 heads, batch size 64)
    on A100 GPU. Fewer HBM accesses result in faster runtime, up to a point.
    \textbf{Right}: The runtime (for seq.\ length 4K) of
  block-sparse \sysname is faster than \sysname by a factor proportional
  to the sparsity.
    }
    \vspace{-1.0em}
\end{figure}

We validate that the number of HBM accesses is the main determining factor of
attention run-time.
In~\cref{fig:micros} (left), we see that even though \sysname has
higher FLOP count compared to standard attention (due to recomputation in the
backward pass), it has much fewer HBM accesses, resulting in much faster
runtime.
In~\cref{fig:micros} (middle), we vary the block size $B_c$ of \sysname, which results in different amounts of HBM accesses, and measure the
runtime of the forward pass.
As block size increases, the number of HBM accesses decreases (as we make fewer
passes over the input), and runtime decreases.
For large enough block size (beyond 256), the runtime is then bottlenecked by
other factors (e.g., arithmetic operations).
Moreover, larger block size will not fit into the small SRAM size.

\subsection{Extension: Block-Sparse \sysname}
\label{sec:blocks_sparse}

We extend \sysname to approximate attention:
we propose block-sparse \sysname, whose IO
complexity is smaller than \sysname by a factor proportional to the
sparsity.

Given inputs $\vQ, \vK, \vV \in \mathbb{R}^{N \times d}$ and a mask matrix
$\tilde{\vM} \in \{ 0, 1 \}^{N \times N}$, we want to compute:
\begin{equation*}
  \vS = \vQ \vK^\top \in \mathbb{R}^{N \times N}, \quad \vP = \softmax(\vS \odot \vmathbb{1}_{\tilde{\vM}}) \in \mathbb{R}^{N \times N}, \quad \vO = \vP\vV \in \mathbb{R}^{N \times d},
\end{equation*}
where $(\vS \odot \vmathbb{1}_{\tilde{\vM}})_{kl} = \vS_{kl}$ if
$\tilde{\vM}_{kl} = 1$ and $-\infty$ if $\vM_{kl} = 0$.
We require $\tilde{\vM}$ to have block form: for some block sizes $B_r, B_c$,
for all $k, l$, $\tilde{\vM}_{k, l} = \vM_{ij}$ with
$i = \lfloor k / B_r \rfloor, j = \lfloor l / B_c \rfloor$ for some $\vM \in \{ 0, 1 \}^{N/B_r \times N/B_c}$.

Given a predefined block sparsity mask $\vM \in \{ 0, 1 \}^{N/B_r \times N/B_c}$ we can
easily adapt~\cref{alg:stream_attn} to only compute the nonzero blocks of the
attention matrix.
The algorithm is identical to~\cref{alg:stream_attn}, except we skip zero
blocks.
We reproduce the algorithm description in~\cref{alg:blocksparse_stream_attn} in
\cref{sec:algo_details}.

We also analyze the IO complexity of block-sparse \sysname.
\begin{proposition}\label{thm:io_complexity_blocksparse}
  Let $N$ be the sequence length, $d$ be the head dimension, and $M$ be size of
  SRAM with $d \leq M \leq Nd$.
  Block-sparse \sysname (\cref{alg:blocksparse_stream_attn}) requires
  $\Theta ( Nd + N^2 d^2 M^{-1} s )$ HBM accesses where $s$ is the
  fraction of nonzero blocks in the block-sparsity mask.
\end{proposition}
We see that applying block-sparsity yields a direct improvement by the sparsity to the larger term in the IO
complexity.
For large sequence lengths $N$, $s$ is often set to
$N^{-1/2}$~\citep{child2019generating} or
$N^{-1}\log N$~\citep{zaheer2020bigbird, beltagy2020longformer, dao2021pixelated}, resulting in $\Theta(N\sqrt{N})$ or
$\Theta(N \log N)$ IO complexity.
For downstream experiments, we use the fixed butterfly sparsity
pattern~\citep{dao2021pixelated}, which has been shown to be able to approximate
arbitrary sparsity~\citep{dao2020kaleidoscope}.

In~\cref{fig:micros} (right), we validate that as the sparsity increases, the
runtime of block-sparse \sysname improves proportionally.
On the LRA benchmark, block-sparse \sysname achieves 2.8$\times$
speedup, while performing on par with standard attention (\cref{sec:exp}).

\section{Experiments}
\label{sec:exp}

We evaluate the impact of using \sysname to train Transformer models.
We validate two claims about training time and model accuracy, and report attention runtime and memory benchmarks.
\begin{itemize}[itemsep=0.1pt,topsep=0pt,leftmargin=*]
    \item \textbf{Training Speed.} \sysname outperforms the MLPerf 1.1~\citep{mattson2020mlperf} speed record for BERT by 15\%, and speeds up GPT-2 up to 3$\times$ over HuggingFace~\citep{wolf-etal-2020-transformers} and $1.8\times$ over Megatron~\citep{shoeybi2019megatron} over standard Transformers.
    \sysname speeds up the long-range
    arena (LRA) benchmark 2.4$\times$. %
    \item \textbf{Quality.} \sysname scales Transformers to longer sequences, yielding higher quality. %
    \sysname trains GPT-2 with context length 4K faster
    than Megatron trains GPT-2 with context length 1K, while
    achieving 0.7 better perplexity.
    Modeling longer sequences yields 6.4 points of lift on two long-document classification tasks.
    Finally, \sysname yields the \textbf{first Transformer} that can achieve
    better-than-random performance on the challenging Path-X task (sequence
    length 16K), and block-sparse \sysname yields the \textbf{first sequence model} that we know of that can achieve better-than-random performance on Path-256 (sequence length 64K).
    \item \textbf{Benchmarking Attention.} We measure the runtime and memory performance of \sysname and block-sparse \sysname based on sequence length.
We confirm that the memory footprint of \sysname scales linearly with seq.\
length and is up to 3$\times$ faster than standard attention for common seq.\
lengths (up to 2K).
We confirm that runtime of block-sparse \sysname scales linearly in seq.\ length and is faster than all existing approximate attention baselines.
\end{itemize}
Additional experiment details are in~\cref{sec:experiment_details}.

\subsection{Faster Models with \sysname}
\label{ssec:exp_language_model}

\paragraph{BERT.}
\sysname yields the fastest single-node BERT training speed that we know of.
We train a BERT-large~\citep{devlin2018bert} model
with \sysname on Wikipedia.
\cref{table:bert_speed} compares our training time to the implementation from Nvidia that set the
training speed record for MLPerf 1.1~\citep{mattson2020mlperf}.
Our implementation is 15\% faster.

\begin{table}[h]
  \captionsetup{font=small}
  \small
  \centering
  \vspace{-1em}
  \caption{\label{table:bert_speed}Training time of BERT-large,
    starting
    from the same initialization provided by the MLPerf benchmark,
    to reach the
    target accuracy of 72.0\% on masked language modeling.
    Averaged over 10 runs on 8$\times$A100 GPUs.
}
  \iftoggle{arxiv}{}{
    \resizebox{0.5\linewidth}{!}
  }
  {
    \begin{tabular}{@{}c|c@{}}
      BERT Implementation & Training time (minutes)  \\ \hline
      Nvidia MLPerf 1.1~\citep{mattson2020mlperf} & 20.0 $\pm$ 1.5 \\
      \sysname (ours) & \textbf{17.4} $\pm$ 1.4 \\
    \end{tabular}
  }
  \vspace{-1em}
\end{table}

\paragraph{GPT-2.}
\sysname yields faster training times for GPT-2~\citep{radford2019language} on the large OpenWebtext dataset~\citep{Gokaslan2019OpenWeb} than the widely used HuggingFace~\citep{wolf-etal-2020-transformers} and Megatron-LM~\citep{shoeybi2019megatron} implementations.
Table~\ref{table:gpt_finetune} shows up to 3$\times$ end-to-end speedup compared to Huggingface
and 1.7$\times$ speedup compared to Megatron-LM.
\sysname achieves the same perplexity as the other two
implementations, as we do not change the model definition.
\cref{sec:experiment_details} includes plots of the validation perplexity throughout training,
confirming that \sysname is as numerically stable as the baselines
and produces the same training / validation curves.

\begin{table}[h]
  \captionsetup{font=small}
  \small
  \centering
  \vspace{-1em}
  \caption{\label{table:gpt_finetune}GPT-2 small and medium using \sysname achieve up to 3$\times$ speed up compared to
    Huggingface implementation and up to 1.7$\times$ compared to Megatron-LM.
    Training time reported on 8$\times$A100s GPUs.}
  \setlength{\tabcolsep}{5pt}
  \iftoggle{arxiv}{}{
      \resizebox{0.8\linewidth}{!}
  }
  {
    \begin{tabular}{@{}c|cc@{}}
      Model implementations &OpenWebText (ppl)& Training time (speedup) \\
      \hline
      GPT-2 small - Huggingface~\citep{wolf-etal-2020-transformers} & 18.2 & 9.5 days (1.0$\times$) \\
      GPT-2 small - Megatron-LM~\citep{shoeybi2019megatron} & 18.2 & 4.7 days (2.0$\times$) \\
      GPT-2 small - \sysname & 18.2 & \textbf{2.7 days (3.5$\times$)} \\ \hline
      GPT-2 medium - Huggingface~\citep{wolf-etal-2020-transformers} & 14.2 & 21.0 days (1.0$\times$) \\
      GPT-2 medium - Megatron-LM~\citep{shoeybi2019megatron} & 14.3 & 11.5 days (1.8$\times$) \\
      GPT-2 medium - \sysname & 14.3 & \textbf{6.9 days (3.0$\times$)} \\
      
    \end{tabular}
  }
  \vspace{-1.5em}
\end{table}

\paragraph{Long-range Arena.}
We compare vanilla Transformer (with either standard implementation or \sysname)
on the long-range arena (LRA~\citep{tay2020long}) benchmark.
We measure accuracy, throughput, and training time of all models.
Each task has a different sequence length varying between 1024 and 4096.
We follow the implementation and experimental setting
in~\citet{tay2020long}and~\citet{xiong2021nystromformer}.\footnote{LRA accuracy
  results are known to be highly dependent on the tuning
  procedure~\citep{xiong2021nystromformer}.
  Our reproduced baselines perform better than as reported in the original
  comparison~\citep{tay2020long}.}
\cref{table:lra} shows that \sysname achieves up 2.4$\times$
speed-up compared to standard attention.
Block-sparse \sysname is faster than all of the approximate attention methods that we have
tested.

\begin{table}[h]
\captionsetup{font=small}
  \vspace{-1em}
    \caption{The performance of standard attention, \sysname, block-sparse
      \sysname, and approximate attention baselines on the Long-Range-Arena benchmarks.}
	\centering
	\small
  \iftoggle{arxiv}{}{
    \resizebox{0.9\linewidth}{!}
  }
  {
	\begin{tabular}{c|ccccc|c|c}
  Models & ListOps & Text & Retrieval & Image & Pathfinder & Avg & Speedup \\
	\hline
	Transformer & 36.0 & 63.6 & 81.6 & 42.3 & 72.7 & 59.3 & - \\
  \sysname & 37.6 & 63.9 & 81.4 & 43.5 & 72.7 & 59.8 & 2.4$\times$ \\
  Block-sparse \sysname & 37.0 & 63.0 & 81.3 & 43.6 & 73.3 & 59.6 & \textbf{2.8$\times$} \\
	\cline{1-8}
	\hline
  Linformer~\citep{wang2020linformer} & 35.6 & 55.9 & 77.7 & 37.8 & 67.6 & 54.9 & 2.5$\times$ \\
  Linear Attention~\citep{katharopoulos2020transformers} & 38.8 & 63.2 & 80.7 & 42.6 & 72.5 & 59.6 & 2.3$\times$ \\
  Performer~\citep{choromanski2020rethinking} & 36.8 & 63.6 & 82.2 & 42.1 & 69.9 & 58.9 & 1.8$\times$ \\
  Local Attention~\citep{tay2020long} & 36.1 & 60.2 & 76.7 & 40.6 & 66.6 & 56.0 & 1.7$\times$ \\
  Reformer~\citep{kitaev2020reformer} & 36.5 & 63.8 & 78.5 & 39.6 & 69.4 & 57.6 & 1.3$\times$  \\
  Smyrf~\citep{daras2020smyrf} & 36.1 & 64.1 & 79.0 & 39.6 & 70.5 & 57.9 & 1.7$\times$ \\
	\end{tabular}
  }
	\label{table:lra}
	\vspace{-1em}
\end{table}

\subsection{Better Models with Longer Sequences}
\label{ssec:exp_long_sequences}

\paragraph{Language Modeling with Long Context.}
The runtime and memory-efficiency of \sysname allow us to increase the context length of
GPT-2 by 4$\times$ while still running faster than the optimized
implementation from Megatron-LM.
\cref{table:gpt2_long_context} shows that that GPT-2 with \sysname and
context length 4K is still 30\% faster than GPT-2 from Megatron with context
length 1K, while achieving 0.7 better perplexity.

\begin{table}[h]
\vspace{-3mm}
  \captionsetup{font=small}
  \small
  \centering
  \caption{\label{table:gpt2_long_context}GPT-2 small with \sysname, with 4$\times$ larger context
    length compared to Megatron-LM, is still 30\% faster while achieving 0.7
    better perplexity. Training time on 8$\times$A100 GPUs is reported.}
  \setlength{\tabcolsep}{5pt}
  \vspace{1em}
  \iftoggle{arxiv}{}{
      \resizebox{0.8\linewidth}{!}
  }
  {
    \begin{tabular}{@{}c|ccc@{}}
      Model implementations & Context length &\multicolumn{1}{c}{OpenWebText (ppl)}&\multicolumn{1}{c}{Training time (speedup)} \\
    \hline
      GPT-2 small - Megatron-LM & 1k & 18.2 & 4.7 days (1.0$\times$) \\
      GPT-2 small - \sysname & 1k & 18.2 & \textbf{2.7 days (1.7$\times$)} \\
      GPT-2 small - \sysname & 2k & 17.6 & 3.0 days (1.6$\times$) \\
      GPT-2 small - \sysname & 4k & \textbf{17.5} & 3.6 days (1.3$\times$) \\
    \end{tabular}
  }
  \vspace{-3mm}
\end{table}

\paragraph{Long Document Classification.}
Training Transformers with longer sequences with \sysname improves performance on the MIMIC-III~\citep{johnson2016mimic} and ECtHR~\citep{chalkidis-etal-2019-neural, chalkidis-et-al-2021-ecthr} datasets.
MIMIC-III contains intensive care unit patient discharge summaries, each annotated with multiple labels.
ECtHR contains legal cases from the European Court of Human Rights, each of which is mapped to articles of the Convention of Human Rights that were allegedly violaged.
Both of these datasets contain very long text documents; the average number of tokens in MIMIC is 2,395 tokens, and the longest document contains 14,562 tokens, while the average and longest numbers in ECtHR are 2,197 and 49,392, respectively.
We evaluate lift from increasing the sequence length of a pretrained RoBERTa model~\citep{liu2019roberta} (we repeat the positional embeddings, as in~\citet{beltagy2020longformer}).

Table~\ref{tab:mimic} shows that sequence length 16K outperforms length 512 by 4.3 points on MIMIC, and that  length 8K outperforms length 512 by 8.5 points on ECtHR.
The discrepancies may be due to subtle distribution shifts: MIMIC-III contains specialized medical text and thus may be more susceptible to a distribution shift in the document length, whereas ECtHR contains general language.

\vspace{-1em}
\begin{table}[h]
    \centering
    \begin{minipage}{2.5in}
    \small
\captionsetup{font=small}
\caption{Long Document performance (micro $F_1$) at different sequence lengths using \sysname.}
\resizebox{1.05\linewidth}{!}
{
\begin{tabular}{@{}r|ccccccccc@{}}
 & 512 & 1024 & 2048 & 4096 & 8192 & 16384 \\
\hline
MIMIC-III~\citep{johnson2016mimic} & 52.8 & 50.7 & 51.7 & 54.6 & 56.4 & \textbf{57.1} \\
ECtHR~\citep{chalkidis-etal-2019-neural} & 72.2 & 74.3 & 77.1 & 78.6 & \textbf{80.7} & 79.2 \\
\end{tabular}
}
\label{tab:mimic}

    \end{minipage}
    \begin{minipage}{0.20in}
    ~
    \end{minipage}
    \begin{minipage}{2.5in}
    \captionsetup{font=small}
    \caption{We report the first Transformer model that can achieve non-random performance on Path-X and Path-256.}
	\centering
	\small
  \resizebox{0.95\linewidth}{!}
  {
	\begin{tabular}{c|cc}
    {\bf Model}  & Path-X & Path-256 \\
	\hline
	Transformer& \xmark & \xmark \\
	Linformer~\citep{wang2020linformer}& \xmark & \xmark \\
	Linear Attention~\citep{katharopoulos2020transformers}& \xmark & \xmark \\
	Performer~\citep{choromanski2020rethinking}& \xmark & \xmark \\
	Local Attention~\citep{tay2020long} & \xmark & \xmark \\
	Reformer~\citep{kitaev2020reformer}& \xmark & \xmark \\
	SMYRF~\citep{daras2020smyrf}& \xmark & \xmark \\
	
    \hline
    \sysname & \textbf{61.4} & \xmark \\
    Block-sparse \sysname & 56.0 & \textbf{63.1} \\
	
	\end{tabular}
  }
	\label{table:pathx}

    \end{minipage}
\end{table}
\vspace{-1em}

\paragraph{Path-X and Path-256.}
The Path-X and Path-256 benchmarks are challenging tasks from the long-range arena benchmark designed to test long context.
The task is to classify whether two points in a black and white 128$\times$128 (or 256$\times$256) image have a path connecting them, and the images are fed to the transformer one pixel at a time.
In prior work, all transformer models have either run out of memory, or only
achieved random performance~\citep{tay2020long}.
There has been a search for alternative architectures that can model such long context~\citep{gu2022efficiently}.
We present here the first result of Transformer models being able to solve
Path-X and Path-256 (\cref{table:pathx}).
We pretrain a transformer on Path-64, and then transfer to Path-X by spatially interpolating the positional embeddings.
\sysname achieves 61.4 accuracy on Path-X.
Additionally, block-sparse \sysname enables the Transformers to scale to sequence length 64K, achieving 63.1 accuracy\footnote{Path-256 requires longer sequences but has relatively shorter paths than Path-X, so it is easier to obtain a higher accuracy.} on Path-256.

\subsection{Benchmarking Attention}
\label{sec:benchmark}

\begin{figure}
\centering
\includegraphics[width=5.5in]{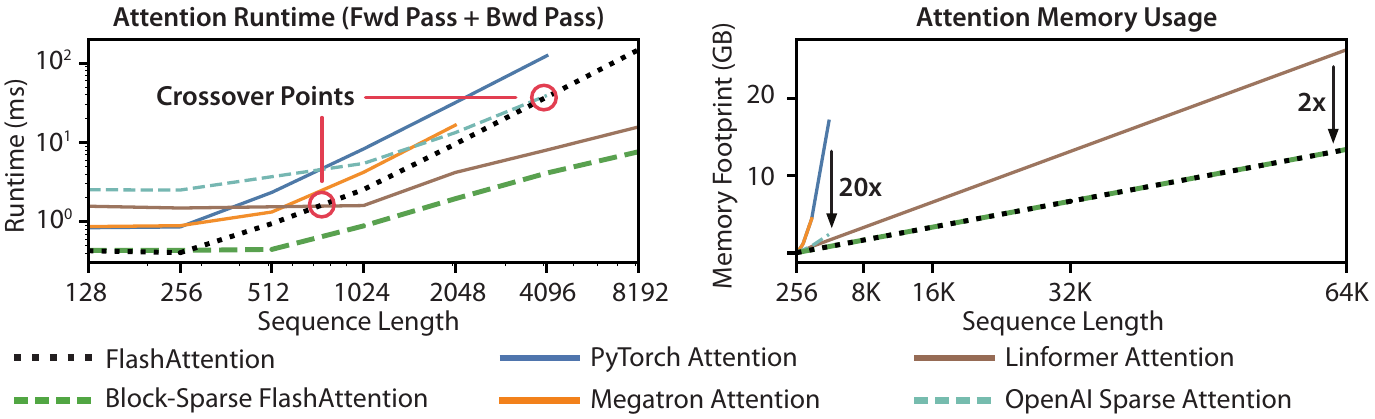}
\iftoggle{arxiv}{}{
\vspace{-1em}
}
\caption{\textbf{Left:} runtime of forward pass + backward pass. \textbf{Right:} attention memory usage.}
\label{fig:benchmark}
\iftoggle{arxiv}{}{
\vspace{-1.0em}
}
\end{figure}

We vary sequence length and measure runtime and memory usage of \sysname and block-sparse \sysname against various attention baselines on one A100 GPU with 40 GB HBM, with dropout and a padding mask.
We compare against reference implementations for exact attention, approximate attention, and sparse attention.
We report a subset of baselines in the main body; Appendix~\ref{sec:experiment_details} contains more baselines and full details.

\paragraph{Runtime.}
Figure~\ref{fig:benchmark} (left) reports the runtime in milliseconds of the forward + backward pass of \sysname and block-sparse \sysname compared to the baselines in exact, approximate, and sparse attention (exact numbers in Appendix~\ref{sec:experiment_details}).
Runtime grows quadratically with sequence length, but \sysname runs significantly faster than \textbf{exact attention} baselines, up to 3$\times$ faster than the PyTorch implementation.
The runtimes of many approximate/sparse attention mechanisms grow linearly with sequence length, but \sysname still runs faster than approximate and sparse attention for short sequences due to fewer memory accesses.
The \textbf{approximate attention} runtimes begin to cross over with \sysname at sequences between 512 and 1024.
On the other hand, block-sparse \sysname is faster than all implementations of exact, sparse, and approximate attention that we know of, across all sequence lengths.

\paragraph{Memory Footprint.}
Figure~\ref{fig:benchmark} (right) shows the memory footprint of \sysname and block-sparse \sysname compared to various exact, approximate, and sparse attention baselines.
\sysname and block-sparse \sysname have the same memory footprint, which grows linearly with sequence length.
\sysname is up to 20$\times$ more memory efficient than \textbf{exact attention} baselines, and is more memory-efficient than the \textbf{approximate attention} baselines.
All other algorithms except for Linformer run out of memory on an A100 GPU before 64K, and \sysname is still 2$\times$ more efficient than Linformer.

\section{Limitations and Future Directions}
\label{sec:discussion}

We discuss limitations of our approach and future directions. Related work is given in~\cref{sec:related_work}.

\textbf{Compiling to CUDA.} Our current approach to building IO-aware implementations of attention requires writing a new CUDA kernel for each new attention implementation.
This requires writing the attention algorithm in a considerably lower-level language than PyTorch, and requires significant engineering effort.
Implementations may also not be transferrable across GPU architectures.
These limitations suggest the need for a method that supports writing attention
algorithms in a high-level language (e.g., PyTorch), and compiling to IO-aware implementations in CUDA---similar to efforts such as Halide in image processing~\citep{ragan2013halide}.

\textbf{IO-Aware Deep Learning.}
We believe that the IO-aware approach can extend beyond attention.
Attention is the most memory-intensive computation in Transformers, but every layer in a deep network touches GPU HBM.
We hope our work inspires IO-aware implementations of additional modules.
We discuss these potential extensions in~\cref{sec:extension_details}.

\textbf{Multi-GPU IO-Aware Methods.}
Our IO-aware implementation of attention is optimal within constants for computing attention on a single GPU.
However, the attention computation may be parallelizable across multiple GPUs~\citep{recht2013parallel}.
Using multiple GPUs adds an additional layer to IO analysis---accounting for data transfer between GPUs.
We hope our work inspires future work in this direction.

\iftoggle{arxiv}{}{
\textbf{Societal Impacts.}
As Transformer-based foundation models grow in size and data, our work seeks to understand how to train these large models more efficiently.
This may allow a general community with limited access to computational resources to train and understand those foundation models.
Our method is applicable to all Transformer-based models, which have a variety of applications, both positive and negative. For example, language modeling may make it easier to spread misinformation, while image classification models may make automatic surveillance easier.
Alleviating these risks requires addressing application-specific issues such as privacy, bias, and discrimination.
}

\subsubsection*{Acknowledgments}

Our implementation uses Apex's FMHA code (\url{https://github.com/NVIDIA/apex/tree/master/apex/contrib/csrc/fmha}) as a starting point.
We thank Young-Jun Ko for the in-depth explanation of his FMHA implementation and for his thoughtful answers to our questions about CUDA.
We thank Sabri Eyuboglu, Megan Leszczynski, Laurel Orr, Yuhuai Wu, Beidi Chen, and Xun Huang for their constructive feedback and suggestions on early drafts of the paper.
We thank Markus Rabe and Charles Staats for helpful discussion of their attention algorithm.

We gratefully acknowledge the support of NIH under No.\ U54EB020405 (Mobilize), NSF under Nos.\ CCF1763315 (Beyond Sparsity), CCF1563078 (Volume to Velocity), and 1937301 (RTML); ARL under No.\ W911NF-21-2-0251 (Interactive Human-AI Teaming); ONR under No.\ N000141712266 (Unifying Weak Supervision); ONR N00014-20-1-2480: Understanding and Applying Non-Euclidean Geometry in Machine Learning; N000142012275 (NEPTUNE); NXP, Xilinx, LETI-CEA, Intel, IBM, Microsoft, NEC, Toshiba, TSMC, ARM, Hitachi, BASF, Accenture, Ericsson, Qualcomm, Analog Devices, Google Cloud, Salesforce, Total, the HAI-GCP \& HAI-Azure Cloud Credits for Research program,  the Stanford Data Science Initiative (SDSI), Department of Defense (DoD) through the National Defense Science and Engineering Graduate Fellowship (NDSEG) Program, and members of the Stanford DAWN project: Facebook, Google, and VMWare. The U.S.\ Government is authorized to reproduce and distribute reprints for Governmental purposes notwithstanding any copyright notation thereon. Any opinions, findings, and conclusions or recommendations expressed in this material are those of the authors and do not necessarily reflect the views, policies, or endorsements, either expressed or implied, of NIH, ONR, or the U.S.\ Government.
Atri Rudra’s research is supported by NSF grant CCF-1763481.

\bibliography{ref}
\bibliographystyle{plainnat}

\newpage

\appendix

\iftoggle{arxiv}{}{
\section*{Checklist}

\begin{enumerate}

\item For all authors...
\begin{enumerate}
  \item Do the main claims made in the abstract and introduction accurately reflect the paper's contributions and scope?
    \answerYes{}
  \item Did you describe the limitations of your work?
    \answerYes{See \cref{sec:discussion}}
  \item Did you discuss any potential negative societal impacts of your work?
    \answerYes{See \cref{sec:discussion}}
  \item Have you read the ethics review guidelines and ensured that your paper conforms to them?
    \answerYes{}
\end{enumerate}

\item If you are including theoretical results...
\begin{enumerate}
  \item Did you state the full set of assumptions of all theoretical results?
    \answerYes{See \cref{sec:theory}}
  \item Did you include complete proofs of all theoretical results?
    \answerYes{See \cref{sec:proofs}}
\end{enumerate}

\item If you ran experiments...
\begin{enumerate}
  \item Did you include the code, data, and instructions needed to reproduce the main experimental results (either in the supplemental material or as a URL)?
    \answerYes{See \cref{sec:experiment_details}}
  \item Did you specify all the training details (e.g., data splits, hyperparameters, how they were chosen)?
    \answerYes{See \cref{sec:experiment_details}}
  \item Did you report error bars (e.g., with respect to the random seed after running experiments multiple times)?
    \answerYes{See \cref{sec:exp}}
  \item Did you include the total amount of compute and the type of resources used (e.g., type of GPUs, internal cluster, or cloud provider)?
    \answerYes{See \cref{sec:experiment_details}}
\end{enumerate}

\item If you are using existing assets (e.g., code, data, models) or curating/releasing new assets...
\begin{enumerate}
  \item If your work uses existing assets, did you cite the creators?
    \answerYes{See \cref{sec:exp,sec:experiment_details}}
  \item Did you mention the license of the assets?
    \answerYes{See \cref{sec:experiment_details}}
  \item Did you include any new assets either in the supplemental material or as a URL?
    \answerNo{}
  \item Did you discuss whether and how consent was obtained from people whose data you're using/curating?
    \answerNA{}
  \item Did you discuss whether the data you are using/curating contains personally identifiable information or offensive content?
    \answerNA{}
\end{enumerate}

\item If you used crowdsourcing or conducted research with human subjects...
\begin{enumerate}
  \item Did you include the full text of instructions given to participants and screenshots, if applicable?
    \answerNA{}
  \item Did you describe any potential participant risks, with links to Institutional Review Board (IRB) approvals, if applicable?
    \answerNA{}
  \item Did you include the estimated hourly wage paid to participants and the total amount spent on participant compensation?
    \answerNA{}
\end{enumerate}

\end{enumerate}

\newpage
}

\section{Related Work}
\label{sec:related_work}

\textbf{IO-Aware Runtime Optimization.}
The broad concept of optimizing for reading and writing to fast/slow memory has a long history in computer science and has been known by many names.
We draw the most direct connection to the literature of analyzing I/O complexity in this work~\citep{aggarwal1988input}, but concepts of memory hierarchies are fundamental and has appeared in many forms, from the working set model~\citep{denning1968working}, to data locality~\citep{wolf1991data}, to the Roofline model of arithmetic intensity~\citep{williams2009roofline}, to analyses of scalability~\citep{mcsherry2015scalability}, to standard textbook treatments of computer architecture~\citep{hennessy2003memory}.
We hope that this work encourages the community to adopt these ideas in more parts of the deep learning stack.

\textbf{Efficient ML Models with Structured Matrices.}
Matrix multiply is the core computational bottleneck of most machine learning
models.
To reduce the computational complexity, there have been numerous approaches to
learn over a more efficient set of matrices.
These matrices are called \emph{structured matrices}, which have subquadratic
($o(n^2)$ for dimension $n \times n$) number of parameters and runtime.
Most common examples of structured matrices are sparse and low-rank matrices,
along with fast transforms commonly encountered in signal processing (Fourier,
Chebyshev, sine/cosine, orthogonal polynomials).
There have been several more general classes of structured matrices proposed in
machine learning: Toeplitz-like~\citep{sindhwani2015structured},
low-displacement rank~\citep{kailath1979displacement},
quasi-separable~\citep{eidelman1999new}).
The butterfly pattern we use for our block-sparse attention is motivated by the
fact that butterfly matrices~\citep{parker1995random, dao2019learning} and their
products have been shown to be able to express any structured matrices with
almost optimal runtime and number of
parameters~\citep{desa2018two,dao2020kaleidoscope}.
However, even though structured matrices are efficient in theory, they have not
seen wide adoption since it is hard to translate their efficiency to wall-clock
speedup since dense unconstrained matrix multiply has very optimize
implementation, a phenomenon known as the hardware
lottery~\citep{hooker2020hardware}.
Extensions of butterfly matrices~\citep{dao2021pixelated,dao2022monarch} aimed
to make butterfly matrices more hardware-friendly.

\textbf{Sparse Training.}
Our block-sparse \sysname can be seen as a step towards making sparse model
training more efficient.
Sparse models have seen success in compressing models for inference (pruning) by
sparsifying the weight
matrices~\citep{han2015deep,han2015learning,sanh2020movement,
  NIPS2017_a51fb975,dong2017learning}.
For model training, the lottery
tickets hypothesis~\citep{frankle2018lottery,frankle2019stabilizing,frankle2020linear}
suggests that there are a set of small sub-networks derived from a larger dense
network that performs as well as the original dense network.
Out block-sparse \sysname can also be seen as a fixed lottery ticket in the
context of attention: we fix the sparsity pattern to be the butterfly pattern
through training, and observe that it performs almost as well as the (dense)
\sysname on the Long-range Arena tasks.

\textbf{Efficient Transformer.}
Transformer-based models have become the most widely-used architecture in
natural language processing~\citep{devlin2018bert} and computer
vision~\citep{dosovitskiy2020image,yuan2021tokens}.
However, one of their computational bottlenecks is that their time and memory
scales quadratic in the sequence length.
There are numerous approaches to overcome this bottleneck, including
approximation with hashing (i.e., sparse) such as
Reformer~\citep{kitaev2020reformer} and Smyrf~\citep{daras2020smyrf} and with
low-rank approximation such as
Performer~\citep{choromanski2020rethinking,likhosherstov2020sub}.
One can even combine sparse and low-rank approximation for better accuracy
(e.g., Longformer~\citep{beltagy2020longformer},
BigBird~\citep{zaheer2020bigbird}, Scatterbrain~\citep{scatterbrain},
Long-short transformer~\citep{zhu2021long}, Combiner~\citep{ren2021combiner}).
Other approaches include compressing along the sequence dimension to attend to
multiple tokens at
once~\citep{wu2019pay,sukhbaatar2019adaptive,lan2019albert,ma2021luna}.
One can also attend over the states from previous sequences to help lengthen the
context (e.g., Transformer-XL~\citep{dai2019transformer} and Compressive
Transformer~\citep{rae2019compressive}).
We recommend the survey~\citep{tay2020efficient} for more details.

There are several lines of work on developing other modules instead of attention
to model longer context. HiPPO~\citep{gu2020hippo} and its extensions, most
notably S4~\citep{gu2021combining, gu2022efficiently, goel2022s} projects the
history on a polynomial basis, allowing accurate reconstruction of the history
through state-space models.
They combine the strengths of CNNs (efficient training), RNNs (efficient
inference), and continuous models (robust to change in sampling rates).
LambdaNetworks~\citep{bello2021lambdanetworks},  AFT~\citep{zhai2021attention}
and FLASH~\citep{hua2022transformer} are other attempts at replacing attention
in the context of image classification and language modeling.

\section{Algorithm Details}
\label{sec:algo_details}

We first derive the forward and backward passes of attention and show that
they can be computed in a memory-efficient manner (requiring extra memory linear
instead of quadratic in the sequence length).
Though they reduce the amount of extra memory required, naively they still incur
quadratic HBM accesses, resulting in slower execution speed.
We describe the \sysname algorithm to implement both the forward and the
backward passes on GPUs that reduces HBM accesses, leading to both faster
runtime and smaller memory footprint.

\subsection{Memory-efficient forward pass}
\label{sec:forward}

The main challenge in making attention memory-efficient is the softmax that
couples the columns of $\vK$ (and columns of $\vV$).
Our approach is to compute the softmax normalization constant separately to
decouple the columns.
This technique~\citep{milakov2018online} has been used in the
literature~\citep{kitaev2020reformer,rabe2021self} to show that attention
computation does not need quadratic \emph{extra} memory (though the number of
HBM accesses is still quadratic, resulting in slow run-time).

For simplicity, we omit here the max-shifting step during softmax.
The full algorithm in~\cref{sec:algo_fwd_full} contains all the steps.

Recall that given input sequences $\vQ, \vK, \vV \in \mathbb{R}^{N \times d}$, we want to
compute the attention output $\vO \in \mathbb{R}^{N \times d}$:
\begin{equation*}
  \vS = \vQ \vK^\top \in \mathbb{R}^{N \times N}, \quad \vP = \softmax(\vS) \in \mathbb{R}^{N \times N}, \quad \vO = \vP\vV \in \mathbb{R}^{N \times d}.
\end{equation*}

We have that $S_{ij} = q_i^T k_j$ where $q_i$ and $k_j$ are the $i$-th and
$j$-th columns of $\vQ$ and $\vK$ respectively.
Define the normalization constants of softmax:
\begin{equation}
  \label{eq:L_i}
  L_i = \sum_{j} e^{q_i^T k_j}.
\end{equation}
Let $v_j$ be the $j$-th column of $\vV$, then the $i$-th columns of the output is
\begin{equation}
  \label{eq:forward_oi}
  o_i = P_{i:} \vV = \sum_{j} P_{ij} v_j = \sum_{j} \frac{e^{q_i^T k_j}}{L_i} v_j.
\end{equation}

We see that once $L_i$ is computed, we can compute $o_i$ without extra memory
by repeatedly summing $\frac{e^{q_i^T k_j}}{L_i} v_j$.
Therefore the forward pass can be computed with $O(n)$ extra memory:
\begin{enumerate}
  \item Compute $L_i$ for all $i$ according to \cref{eq:L_i}, which takes $O(n)$
  extra memory.
  \item Compute $o_i$ for all $i$ according to \cref{eq:forward_oi}, which takes
  $O(d)$ extra memory.
\end{enumerate}

\subsection{Memory-efficient backward pass}
\label{sec:backward}

We derive the backward pass of attention and show that it can also be computed
with linear memory.
\citet{rabe2021self} suggests that the backward pass can be done without
quadratic extra memory by applying gradient checkpointing to the
memory-efficient forward pass.
We instead derive the backward pass explicitly and show how it can be computed
in a memory-efficient manner.

Suppose that there is a scalar loss function $\phi$, and let the output gradient
be $\vdO \in \mathbb{R}^{n \times d}$ (where $\vdO$ denotes
$\frac{\partial \phi}{\partial \vO}$).
We want to compute the input gradients $\vdQ, \vdK, \vdV \in \mathbb{R}^{n \times d}$
(where $\vdQ, \vdK, \vdV$ denote
$\frac{\partial \phi}{\partial \vQ}, \frac{\partial \phi}{\partial \vK}, \frac{\partial \phi}{\partial \vV}$
respectively).

The gradient $\vdV$ is easy to see.
Applying reverse-mode autodiff by hand (aka the chain rule), we obtain (in
matrix notation) $\vdV = \vP^T \vdO$.
Thus:
\begin{equation}
  \label{eq:dv}
  dv_j = \sum_{i} P_{ij} do_i = \sum_{i} \frac{e^{q_i^T k_j}}{L_i} do_i.
\end{equation}
Since we already computed $L_i$, $dv_j$ can be computed without extra memory by
repeated summing.

The gradients $\vdQ$ and $\vdK$ are a little more complicated.
We go through the gradients $\vdP$ and $\vdS$ first.
From \cref{eq:forward_oi}, we have that $\vdP = \vdO \vV^T$, and so:
\begin{equation*}
  dP_{ij} = do_i^T v_j.
\end{equation*}

Recall that $P_{i:} = \softmax(S_{i:})$.
Using the fact that the Jacobian of $y = \softmax(x)$ is $\diag(y) - y y^T$, we
have that
\begin{equation*}
  dS_{i:} = (\diag(P_{i:}) - P_{i:} P_{i:}^T) dP_{i:} = P_{i:} \circ dP_{i:} - (P_{i:}^T dP_{i:}) P_{i:},
\end{equation*}
where $\circ$ denotes pointwise multiplication.

Define
\begin{equation}
  \label{eq:D_i}
  D_{i} = P_{i:}^T dP_{i:} = \sum_{j} \frac{e^{q_i^T k_j}}{L_i} do_i^T v_j = do_i^T \sum_{j} \frac{e^{q_i^\top k_j}}{L_i} v_j = do_i^T o_i,
\end{equation}
then
\begin{equation*}
  dS_{i:} = P_{i:} \circ dP_{i:} - D_i P_{i:}.
\end{equation*}
Hence
\begin{equation*}
  dS_{ij} = P_{ij} dP_{ij} - D_i P_{ij} = P_{ij} (dP_{ij} - D_i).
\end{equation*}

Now we can get the gradients $\vdQ$ and $\vdK$.
Recall that $S_{ij} = q_i^T k_j$, so
\begin{equation}
  \label{eq:dq}
  dq_i = \sum_{j} dS_{ij} k_j = \sum_{j} P_{ij} (dP_{ij} - D_i) k_j = \sum_{j} \frac{e^{q_i^T k_j}}{L_i} (do_i^T v_j - D_i) k_j.
\end{equation}
Similarly,
\begin{equation}
  \label{eq:dk}
  dk_j = \sum_{i} dS_{ij} q_i = \sum_{i} P_{ij} (dP_{ij} - D_i) q_i = \sum_{i} \frac{e^{q_i^T k_j}}{L_i} (do_i^T v_j - D_i) q_i.
\end{equation}

Therefore the backward pass can also be computed with $O(n)$ extra memory:
\begin{enumerate}
  \item Compute $dv_j$ for all $j$ according to \cref{eq:dv}, which takes
  $O(d)$ extra memory.
  \item Compute $D_i$ for all $i$ according to \cref{eq:D_i}, which takes $O(n)$
  extra memory.
  \item Compute $dq_i$ for all $i$ according to \cref{eq:dq}, which takes
  $O(d)$ extra memory.
  \item Compute $dk_j$ for all $j$ according to \cref{eq:dk}, which takes
  $O(d)$ extra memory.
\end{enumerate}

\subsection{\sysname: Forward Pass}
\label{sec:algo_fwd_full}

We describe the full details of \sysname forward pass.
Given input sequences $\vQ, \vK, \vV \in \mathbb{R}^{N \times d}$, we want to
compute the attention output $\vO \in \mathbb{R}^{N \times d}$:
\begin{align*}
  &\vS = \tau \vQ \vK^\top \in \mathbb{R}^{N \times N}, \quad
  \vS^{\mathrm{masked}} = \textsc{mask}(S) \in \mathbb{R}^{N \times N}, \quad
  \vP = \softmax(\vS^{\mathrm{masked}}) \in \mathbb{R}^{N \times N}, \\
  &\vP^{\mathrm{dropped}} = \mathrm{dropout}(\vP, p_\mathrm{drop}), \quad
  \vO = \vP^{\mathrm{dropped}}\vV \in \mathbb{R}^{N \times d},
\end{align*}
where $\tau \in \mathbb{R}$ is some softmax scaling (typically $\frac{1}{\sqrt{d}}$),
$\textsc{mask}$ is some masking function that sets some entries of the input to
$-\infty$ and keep other entries the same (e.g., key padding mask when sequences
in the batch don't have the same lengths and are padded), and
$\mathrm{dropout}(x, p)$ applies dropout to $x$ elementwise (i.e., output $\frac{x}{1 - p}$
with probability $1 - p$ and output 0 with probability $p$ for each element $x$).

\iftoggle{icmlworkshop}{
For numerical stability, the softmax of vector $x \in \mathbb{R}^{B}$ is computed as:
\begin{equation*}
  m(x) \defeq \max_i\ \ x_i, \quad
  f(x) \defeq \begin{bmatrix} e^{x_1 - m(x)} & \hdots & e^{x_B - m(x)} \end{bmatrix}, \quad
  \ell(x) \defeq \sum_i f(x)_i, \quad
  \softmax(x) \defeq \frac{f(x)}{\ell(x)}.
\end{equation*}
For vectors $x^{(1)}, x^{(2)} \in \mathbb{R}^{B}$, we can decompose the softmax of the concatenated $x = \begin{bmatrix} x^{(1)} \ x^{(2)} \end{bmatrix} \in \mathbb{R}^{2B}$ as:
\begin{align*}
  &m(x) = m(\begin{bmatrix} x^{(1)} \ x^{(2)} \end{bmatrix}) = \max(m(x^{(1)}), m(x^{(2)})), \quad
  f(x) = \begin{bmatrix} e^{m(x^{(1)}) - m(x)} f(x^{(1)}) & e^{m(x^{(2)}) - m(x)} f(x^{(2)}) \end{bmatrix}, \\
  &\ell(x) = \ell(\begin{bmatrix} x^{(1)} \ x^{(2)} \end{bmatrix}) = e^{m(x^{(1)}) - m(x)}\ell (x^{(1)}) + e^{m(x^{(2)}) - m(x)} \ell(x^{(2)}), \quad
  \softmax(x) = \frac{f(x)}{\ell(x)}.
\end{align*}
}{}

The full algorithm is in~\cref{alg:fwd_full}.
We save the output $\vO$, the softmax statistics $\ell$ and $m$, and the pseudo-random
number generator state ${\cal R}$ for the backward pass.
\begin{algorithm}[H]
  \caption{\small\label{alg:fwd_full}\sysname Forward Pass}
  \begin{algorithmic}[1]
    \REQUIRE Matrices $\vQ, \vK, \vV \in \mathbb{R}^{N \times d}$ in HBM, on-chip SRAM of
    size $M$, softmax scaling constant $\tau \in \mathbb{R}$, masking function
    $\textsc{mask}$, dropout probability $p_\mathrm{drop}$.
    \STATE Initialize the pseudo-random number generator state ${\cal R}$ and save to HBM.
    \STATE Set block sizes $B_c = \left\lceil \frac{M}{4d} \right\rceil, B_r = \min \left( \left\lceil \frac{M}{4d} \right\rceil , d \right)$.
    \STATE Initialize $\vO = (0)_{N \times d} \in \mathbb{R}^{N \times d}, \ell = (0)_N \in \mathbb{R}^{N}, m = (-\infty)_N \in \mathbb{R}^{N}$ in HBM.
    \STATE Divide $\vQ$ into $T_r = \left\lceil\frac{N}{B_r} \right\rceil$ blocks $\vQ_1, \dots, \vQ_{T_r}$ of size $B_r \times d$ each,
    and divide $\vK, \vV$ in to $T_c = \left\lceil \frac{N}{B_c} \right\rceil$ blocks $\vK_1, \dots, \vK_{T_c}$ and
    $\vV_1, \dots, \vV_{T_c}$, of size $B_c \times d$ each.
    \STATE Divide $\vO$ into $T_r$ blocks $\vO_i, \dots, \vO_{T_r}$ of size
    $B_r \times d$ each, divide $\ell$ into $T_r$ blocks $\ell_i, \dots, \ell_{T_r}$ of size
    $B_r$ each, divide $m$ into $T_r$ blocks $m_1, \dots, m_{T_r}$ of size $B_r$ each.
    \FOR{$1 \le j \le T_c$}
      \STATE Load $\vK_j, \vV_j$ from HBM to on-chip SRAM.
      \FOR{$1 \le i \le T_r$}
        \STATE Load $\vQ_i, \vO_i, \ell_i, m_i$ from HBM to on-chip SRAM.
        \STATE On chip, compute $\vS_{ij} = \tau \vQ_i \vK_j^T \in \mathbb{R}^{B_r \times B_c}$.
        \STATE On chip, compute $\vS_{ij}^{\mathrm{masked}} = \textsc{mask}(\vS_{ij})$.
        \STATE On chip, compute $\tilde{m}_{ij} = \mathrm{rowmax}(\vS_{ij}^{\mathrm{masked}}) \in \mathbb{R}^{B_r}$, $\tilde{\vP}_{ij} = \exp(\vS_{ij}^{\mathrm{masked}} - \tilde{m}_{ij}) \in \mathbb{R}^{B_r \times B_c}$ (pointwise),
        $\tilde{\ell}_{ij} = \mathrm{row sum}(\tilde{\vP}_{ij}) \in \mathbb{R}^{B_r}$.
        \STATE On chip, compute $m_i^{\mathrm{new}} = \max(m_i, \tilde{m}_{ij}) \in \mathbb{R}^{B_r}$, $\ell_i^{\mathrm{new}} = e^{m_i - m_i^{\mathrm{new}}} \ell_i + e^{\tilde{m}_{ij} - m_i^{\mathrm{new}}} \tilde{\ell}_{ij} \in \mathbb{R}^{B_r}$.
        \STATE On chip, compute $\tilde{\vP}_{ij}^{\mathrm{dropped}} = \mathrm{dropout}(\tilde{\vP}_{ij}, p_\mathrm{drop})$.
        \STATE Write $\vO_i \leftarrow \diag(\ell_i^{\mathrm{new}})^{-1}(\diag(\ell_i) e^{m_i - m_i^{\mathrm{new}}} \vO_i + e^{\tilde{m}_{ij} - m_i^{\mathrm{new}}}\tilde{\vP}_{ij}^{\mathrm{dropped}} \vV_j)$
        to HBM.
        \STATE Write $\ell_i \leftarrow \ell_i^{\mathrm{new}}$, $m_i \leftarrow m_i^{\mathrm{new}}$ to HBM.
      \ENDFOR
    \ENDFOR
    \STATE Return $\vO, \ell, m, {\cal R}$.
  \end{algorithmic}
\end{algorithm}

\subsection{\sysname: Backward Pass}
\label{sec:algo_bwd_full}

We describe the full details of \sysname backward pass.
Given input sequences $\vQ, \vK, \vV \in \mathbb{R}^{N \times d}$, the output $\vO \in \mathbb{R}^{N \times d}$,
and the output gradient $\vdO$, we want to
compute the input gradients $\vdQ, \vdK, \vdV \in \mathbb{R}^{N \times d}$.

We first describe the standard attention backward pass in~\cref{alg:standard_attn_bwd} for completeness.
\begin{algorithm}[H]
  \caption{\small\label{alg:standard_attn_bwd}Standard Attention Backward Pass}
  \begin{algorithmic}[1]
    \REQUIRE Matrices $\vQ, \vK, \vV, \vdO \in \mathbb{R}^{N \times d}$, $\vP \in \mathbb{R}^{N \times N}$ in HBM.
    \STATE Load $\vP, \vdO$ by blocks from HBM, compute
    $\vdV = \vP^\top \vdO \in \mathbb{R}^{N \times d}$, write $\vdV$ to HBM.
    \STATE Load $\vdO, \vV$ by blocks from HBM, compute
    $\vdP = \vdO \vV^\top \in \mathbb{R}^{N \times N}$, write $\vdP$ to HBM.
    \STATE Read $\vP, \vdP$ from HBM, compute $\vdS \in \mathbb{R}^{N \times N}$ where
    $dS_{ij} = P_{ij} (dP_{ij} - \sum_l P_{il} dP_{il})$, write $\vdS$ to
    HBM.
    \STATE Load $\vdS$ and $\vK$ by blocks from HBM, compute $\vdQ = \vdS\vK$,
    write $\vdQ$ to HBM.
    \STATE Load $\vdS$ and $\vQ$ by blocks from HBM, compute $\vdK = \vdS^\top\vQ$, write $\vdK$ to
    HBM.
    \STATE Return $\vdQ, \vdK, \vdV$.
  \end{algorithmic}
\end{algorithm}

We now make two observations about \sysname backward pass:
\begin{enumerate}
  \item We do not need to store the dropout mask of size $O(N^2)$ from the
  forward pass.
  Instead, we can save the pseudo-random number generator states from
  the forward pass and re-generate the dropout mask in the backward pass.
  This allows us to only use $O(N)$ extra memory.
  \item When computing the softmax gradient, we use~\cref{eq:D_i} to compute
  $D_i = P_{i:}^\top dP_{i:}$ without reducing over $P_{i:}$ and $dP_{i:}$ of size
  $N$ (they might not fit into SRAM).
  Instead we can rewrite $D_i = do_i^\top o_i$ and compute the dot product between
  vectors of size $d$.
\end{enumerate}

The full \sysname backward pass algorithm is in~\cref{alg:bwd_full}.
Conceptually it is just a block version of the derivation
in~\cref{sec:backward}.

\iftoggle{arxiv}{
\begin{algorithm}[H]
} {
\begin{algorithm}[h]
}
  \caption{\small\label{alg:bwd_full}\sysname Backward Pass}
  \begin{algorithmic}[1]
    \REQUIRE Matrices $\vQ, \vK, \vV, \vO, \vdO \in \mathbb{R}^{N \times d}$ in HBM,
    vectors $\ell, m \in \mathbb{R}^N$ in HBM, on-chip SRAM of
    size $M$, softmax scaling constant $\tau \in \mathbb{R}$, masking function
    $\textsc{mask}$, dropout probability $p_\mathrm{drop}$, pseudo-random number
    generator state ${\cal R}$ from the forward pass.
    \STATE Set the pseudo-random number generator state to ${\cal R}$.
    \STATE Set block sizes $B_c = \left\lceil \frac{M}{4d} \right\rceil, B_r = \min \left( \left\lceil \frac{M}{4d} \right\rceil , d \right)$.
    \STATE Divide $\vQ$ into $T_r = \left\lceil\frac{N}{B_r} \right\rceil$ blocks $\vQ_1, \dots, \vQ_{T_r}$ of size $B_r \times d$ each,
    and divide $\vK, \vV$ in to $T_c = \left\lceil \frac{N}{B_c} \right\rceil$ blocks $\vK_1, \dots, \vK_{T_c}$ and
    $\vV_1, \dots, \vV_{T_c}$, of size $B_c \times d$ each.
    \STATE Divide $\vO$ into $T_r$ blocks $\vO_i, \dots, \vO_{T_r}$ of size
    $B_r \times d$ each, divide $\vdO$ into $T_r$ blocks $\vdO_i, \dots, \vdO_{T_r}$
    of size $B_r \times d$ each, divide $\ell$ into $T_r$ blocks $\ell_i, \dots, \ell_{T_r}$ of size
    $B_r$ each, divide $m$ into $T_r$ blocks $m_1, \dots, m_{T_r}$ of size $B_r$ each.
    \STATE Initialize $\vdQ = (0)_{N \times d}$ in HBM and divide it into $T_r$ blocks $\vdQ_1, \dots, \vdQ_{T_r}$ of size $B_r \times d$ each.
    Initialize $\vdK = (0)_{N \times d}, \vdV = (0)_{N \times d}$ in HBM and divide $\vdK, \vdV$ in to $T_c$ blocks $\vdK_1, \dots, \vdK_{T_c}$ and
    $\vdV_1, \dots, \vdV_{T_c}$, of size $B_c \times d$ each.
    \FOR{$1 \le j \le T_c$}
      \STATE Load $\vK_j, \vV_j$ from HBM to on-chip SRAM.
      \STATE Initialize $\tilde{\vdK}_j = (0)_{B_c \times d}, \tilde{\vdV}_j = (0)_{B_c \times d}$ on SRAM.
      \FOR{$1 \le i \le T_r$}
        \STATE Load $\vQ_i, \vO_i, \vdO_i, \vdQ_i, \ell_i, m_i$ from HBM to on-chip SRAM.
        \STATE On chip, compute $\vS_{ij} = \tau \vQ_i \vK_j^T \in \mathbb{R}^{B_r \times B_c}$.
        \STATE On chip, compute $\vS_{ij}^{\mathrm{masked}} = \textsc{mask}(\vS_{ij})$.
        \STATE On chip, compute $\vP_{ij} = \diag(l_i)^{-1}\exp(\vS_{ij}^{\mathrm{masked}} - m_{i}) \in \mathbb{R}^{B_r \times B_c}$.
        \STATE On chip, compute dropout mask $\vZ_{ij} \in \mathbb{R}^{B_r \times B_c}$ where
        each entry has value $\frac{1}{1 - p_{\mathrm{drop}}}$ with probability
          $1 - p_\mathrm{drop}$ and value 0 with probability $p_\mathrm{drop}$.
        \STATE On chip, compute
        $\vP_{ij}^{\mathrm{dropped}} = \vP_{ij} \circ \vZ_{ij}$ (pointwise multiply).
        \STATE On chip, compute
        $\tilde{\vdV_j} \leftarrow \tilde{\vdV_j} + (\vP_{ij}^{\mathrm{dropped}})^\top \vdO_i \in \mathbb{R}^{B_c \times d}$.
        \STATE On chip, compute
        $\vdP_{ij}^{\mathrm{dropped}} = \vdO_{i} \vV_j^\top \in \mathbb{R}^{B_r \times B_c}$.
        \STATE On chip, compute
        $\vdP_{ij} = \vdP_{ij}^{\mathrm{dropped}} \circ \vZ_{ij}$ (pointwise multiply).
        \STATE On chip, compute $D_{i} = \mathrm{rowsum}(\vdO_i \circ \vO_i) \in \mathbb{R}^{B_r}$.
        \STATE On chip, compute $\vdS_{ij} = \vP_{ij} \circ (\vdP_{ij} - D_i) \in \mathbb{R}^{B_r \times B_c}$.
        \STATE Write
        $\vdQ_{i} \leftarrow \vdQ_i + \tau \vdS_{ij} \vK_j \in \mathbb{R}^{B_r \times d}$ to HBM.
        \STATE On chip, compute $\tilde{\vdK}_{j} \leftarrow \tilde{\vdK}_j + \tau \vdS_{ij}^\top \vQ_i \in \mathbb{R}^{B_c \times d}$.
      \ENDFOR
      \STATE Write $\vdK_j \leftarrow \tilde{\vdK_j}, \vdV_j \leftarrow \tilde{\vdV_j}$ to HBM.
    \ENDFOR
    \STATE Return $\vdQ, \vdK, \vdV$.
  \end{algorithmic}
\end{algorithm}

We see that similar to the forward pass, the backward pass performs $O(N^2)$
FLOPs and only requires $O(N)$ extra memory beyond inputs, output, output
gradient, and input gradients.

We analyze the IO-complexity of the backward pass, similar to the forward pass (\cref{thm:io_complexity}).
\begin{theorem}\label{thm:io_complexity_bwd}
  Let $N$ be the sequence length, $d$ be the head dimension, and $M$ be size of
  SRAM with $d \leq M \leq Nd$.
  Standard attention (\cref{alg:standard_attn}) backward pass requires $\Theta(Nd + N^2)$ HBM
  accesses, while \sysname backward pass (\cref{alg:bwd_full}) requires
  $\Theta ( N^2 d^2 M^{-1} )$ HBM accesses.
\end{theorem}
The proof is in~\cref{sec:proofs}.

\subsection{Comparison with \citet{rabe2021self}}
\label{subsec:rabe_comparison}

We describe here some similarities and differences between our \sysname
algorithm and the algorithm of \citet{rabe2021self}.

Conceptually, both \sysname and \citet{rabe2021self} operate on blocks of the
attention matrix using the well-established technique of tiling (or softmax
scaling)~\citep{milakov2018online, kitaev2020reformer}.
To reduce the memory footprint, both methods avoid storing the large attention
matrix in the forward pass and recompute it in the backward pass.

The first major difference is that \citet{rabe2021self} focuses on the reducing
the total memory footprint (maximum amount of GPU memory required) while
\sysname focuses on reducing memory accesses (the number of memory
reads/writes).
As mentioned in~\cref{sec:background}, the amount of memory access is the
primary determining factor of runtime.
Reducing memory accesses also necessarily reduces the total amount of memory
required (e.g., if an operation incurs $A$ memory accesses, then its total
memory requirement is at most $A$).
As a result, \sysname is faster than standard attention (2-4$\times$) while
\citet{rabe2021self} is around the same speed or slightly slower than standard
attention.
In terms of total memory required, both methods offer substantial memory saving.

The second difference between the two methods is the way information is summarized
from each block to pass to the next block.
\citet{rabe2021self} summarizes each block with its temporary output along with the
softmax normalization statistics.
At the end of the forward pass, the temporary outputs of all the blocks are combined using
the statistics to produce the final output.
\sysname instead incrementally updates the output (\cref{alg:stream_attn} line
\ref{alg:stream_attn_aggregate}) after processing each block, so only one copy
of the output is needed (instead of $K$ copies for $K$ blocks).
This means that \sysname has smaller total memory requirement compared to \citet{rabe2021self}.

The final major difference is the way the backward pass is computed.
\citet{rabe2021self} uses gradient checkpointing to recompute the attention
matrix and the temporary output of each block.
\sysname instead simplifies the backward pass analytically (\cref{sec:backward,sec:algo_bwd_full}).
It only recomputes the attention matrix and does not recompute the
temporary output of each block.
This reduces the memory requirement for the backward pass and yields speedup.

\section{Proofs}
\label{sec:proofs}

\begin{proof}[Proof of \cref{thm:correctness}]
  We first count the number of FLOPs and extra memory required.

  The dominating FLOPs are from matrix multiplication.
  In the inner loop, (\cref{alg:stream_attn} line \ref{alg:stream_attn_qk}), we
  compute $\vQ_i \vK_j^\top \in \mathbb{R}^{B_r \times B_c}$ for $\vQ_i \in \mathbb{R}^{B_r \times d}$ and
  $\vK_j \in \mathbb{R}^{B_c \times d}$, which takes $O(B_r B_c d)$ FLOPs.
  We also compute (\cref{alg:stream_attn} line \ref{alg:stream_attn_aggregate}) $\tilde{\vP}_{ij} \vV_j \in \mathbb{R}^{B_r \times d}$ for
  $\tilde{\vP}_{ij} \in \mathbb{R}^{B_r \times B_c}$ and $\vV_j \in \mathbb{R}^{B_c \times d}$, which takes $O(B_r B_c d)$ FLOPs.
  We execute the inner loops
  $T_c T_r = \left\lceil  \frac{N}{B_c} \right\rceil \left\lceil \frac{N}{B_r} \right\rceil$ times.
  Therefore the total number of FLOPs is
  \begin{equation*}
    O \left( \frac{N^2}{B_c B_r} B_r B_c d \right) = O(N^2d).
  \end{equation*}

  In terms of extra memory required, we see that we need $O(N)$ memory to store
  the statistics $(\ell, m)$.

  We now prove the algorithm's correctness by induction on $j$ for
  $0 \leq j \leq T_c$.
  Let $\vK_{:j} \in \mathbb{R}^{jB_c \times d}$ be the first $jB_c$ rows of $\vK$, and similarly
  $\vV_{:j} \in \mathbb{R}^{jB_c \times d}$ the the first $jB_c$ rows of $\vV$.
  Let $\vS_{:, :j} = \vQ \vK_{:j}^\top \in \mathbb{R}^{N \times jB_c}$, and
  $\vP_{:, :j} = \mathrm{softmax}(\vS_{:, :j}) \in \mathbb{R}^{N \times jB_c}$ (softmax applied row-wise).
  Let $m^{j}, \ell^{(j)}, \vO^{(j)}$ be the values of $m, \ell, \vO$ in HBM after the
  $j$-th iteration of the outer loop (\cref{alg:stream_attn} line \ref{alg:stream_attn_outer_loop}).
  (Note that these values of $m, \ell, \vO$ are updated after each iteration of the outer loop.)
  We want to show that after the $j$-th iteration of the outer loop, we have
  computed in HBM:
  \begin{equation*}
    m^{(j)} = \mathrm{rowmax}(\vS_{:, :j}) \in \mathbb{R}^N, \quad
    \ell^{(j)} = \mathrm{rowsum}(\exp(\vS_{:, :j} - m^{(j)})) \in \mathbb{R}^N, \quad
    \vO^{(j)} = \vP_{:, :j} \vV_{:j} \in \mathbb{R}^{N \times d}.
  \end{equation*}

  Based on our initialization (\cref{alg:stream_attn} line
  \ref{alg:stream_attn_init}), this claim is true for $j = 0$ (i.e., before the
  any iteration of the outer loop is executed).
  Suppose that the claim holds for some $j = 0, \dots, T_c - 1$.
  We want to show that the claim also holds for $j + 1$.
  Indeed, when we update the statistics in the inner loop
  (\cref{alg:stream_attn} line \ref{alg:stream_attn_statistics}) on the
  $(j + 1)$-th iteration of the outer loop,
  we update $m^{(j + 1)} = \max(m^{(j)}, \tilde{m})$ where $\tilde{m} \in \mathbb{R}^N$ is the
  row-max of $\vS_{:, j:j+1}$, the slice of $\vS$ from column $jB_c$ to column
  $(j+1)B_c - 1$.
  This implies that
  \begin{equation*}
    m^{(j+1)} = \mathrm{rowmax}(\vS_{:, :j+1}) \in \mathbb{R}^N.
  \end{equation*}
  Similarly, we update
  \begin{equation*}
    \ell^{(j + 1)} = e^{m^{(j)} - m^{(j+1)}} \ell^{(j)} + e^{\tilde{m} - m^{(j+1)}} \tilde{\ell} ,
  \end{equation*}
  where $\tilde{\ell} = \mathrm{rowsum}(\exp(\vS_{:, j:j+1} - \tilde{m})) \in \mathbb{R}^N$.
  By the same algebraic manipulation in~\cref{sec:implementation}, we obtain:
  \begin{equation*}
    \ell^{(j+1)} = \mathrm{rowsum}(\exp(\vS_{:, :j+1} - m^{(j+1)})) \in \mathbb{R}^N.
  \end{equation*}

  Let $\vV_{j:j+1}$ be the slice of $\vV$ from column $jB_c$ to column $(j+1)B_c - 1$,
  we also update:
  \begin{align*}
    \vO^{(j + 1)}
    &= \diag(\ell^{(j+1)})^{-1} (\diag(\ell^{(j)})e^{m^{(j)} - m^{(j+1)}} \vO^{(j)} + e^{\tilde{m} - m^{(j+1)}} \exp(\vS_{j:j+1} - \tilde{m}) \vV_{j:j+1} ) \\
    &= \diag(\ell^{(j+1)})^{-1} (\diag(\ell^{(j)})e^{m^{(j)} - m^{(j+1)}} \vP_{:, :j} \vV_{:j} + e^{-m^{(j+1)}} \exp(\vS_{j:j+1}) \vV_{j:j+1} ) \\
    &= \diag(\ell^{(j+1)})^{-1} (\diag(\ell^{(j)})e^{m^{(j)} - m^{(j+1)}} \diag(\ell^{(j)}) \exp(\vS_{:, :j} - m^{(j)}) \vV_{:j} + e^{-m^{(j+1)}} \exp(\vS_{j:j+1}) \vV_{j:j+1} ) \\
    &= \diag(\ell^{(j+1)})^{-1} (e^{- m^{(j+1)}} \exp(\vS_{:, :j}) \vV_{:j} + e^{-m^{(j+1)}} \exp(\vS_{j:j+1}) \vV_{j:j+1} ) \\
    &= \diag(\ell^{(j+1)})^{-1} (\exp(\vS_{:, :j} - m^{(j+1)}) \vV_{:j} + \exp(\vS_{j:j+1} - m^{(j+1)}) \vV_{j:j+1} ) \\
    &= \diag(\ell^{(j+1)})^{-1} \left( \exp \left( \begin{bmatrix} \vS_{:, :j} & \vS_{j:j+1} \end{bmatrix} - m^{(j+1)} \right) \right) \begin{bmatrix} \vV_{:j} \\ \vV_{j:j+1} \end{bmatrix} \\
    &= \softmax(\vS_{:j+1}) \vV_{:j+1}.
  \end{align*}
  We then see that the claim is also true for $j + 1$.
  By induction, the claim is true for all $j = 0, \dots, T_c$.

  When $j = T_c$, we conclude that the final value of $\vO$ in HBM is
  $\softmax(\vS) \vV = \softmax(\vQ \vK^\top) \vV$.

\end{proof}

\begin{proof}[Proof of \cref{thm:io_complexity}]
  We first analyze the IO complexity of standard attention implementation.
  The inputs $\vQ, \vK, \vV \in \mathbb{R}^{N \times d}$ reside in HBM, and
  the at the end of the algorithm the output $\vO \in \mathbb{R}^{N \times d}$ is
  written to HBM.

  In the first step of computing the matrix multiply $\vS = \vQ \vK^\top$, the inputs $\vQ, \vK$
  are read from HBM and the output $\vS \in \mathbb{R}^{N \times N}$ is
  written to HBM (\cref{alg:standard_attn} line~\ref{alg:standard_attn_qk}).
  This incurs $\Theta(Nd + N^2)$ HBM accesses.

  In the second step of computing $\vP = \softmax(\vS)$, the input $\vS$ is read from
  HBM and the output $\vP$ is written to HBM (\cref{alg:standard_attn} line~\ref{alg:standard_attn_sp}).
  This incurs $\Theta(N^2)$ HBM accesses.

  In the last step of computing $\vO = \vP\vV$, the inputs $\vP, \vV$ are read from global
  memory and the output $\vO$ is written to HBM (\cref{alg:standard_attn} line~\ref{alg:standard_attn_pv}).
  This incurs $\Theta(Nd + N^2)$ HBM accesses.

  Overall, standard attention implementation requires $\Theta(Nd + N^2)$ global
  memory accesses.

  We now analyze the IO complexity of streaming attention.

  Following~\cref{alg:stream_attn}, we see that each element of $\vK$ and $\vV$ is
  loaded from HBM once (\cref{alg:stream_attn} line~\ref{alg:stream_attn_load_kv}).
  We make $T_c$ passes over $\vQ$ and $\vO$, each pass loading all of $\vQ$ and all of
  $\vO$ to HBM (\cref{alg:stream_attn} line~\ref{alg:stream_attn_load_qo}).
  Therefore the number of HBM accesses is
  $\Theta \left( Nd + Nd T_c \right) = \Theta(Nd T_c)$.

  We derive the conditions on the block sizes $B_c$ and $B_r$.
  We need the blocks $\vK_j$ and $\vV_j$ of size $B_c \times d$ to fit into
  on-chip memory, which translates to:
  \begin{equation*}
    B_c d = O(M) \Leftrightarrow B_c = O \left( \frac{M}{d} \right).
  \end{equation*}
  Similarly, we need the blocks $\vQ_i, \vO_i$ of size $B_r \times d$ to fit
  into on-chip memory, which translates to:
  \begin{equation*}
    B_r d = O(M) \Leftrightarrow B_r = O \left( \frac{M}{d} \right).
  \end{equation*}
  Finally, we need the block $\vS_{ij}$ of size $B_r \times  B_c$ to
  fit into on-chip memory, which translates to:
  \begin{equation*}
    B_r B_c = O(M).
  \end{equation*}
  We therefore set:
  \begin{equation*}
    B_c = \Theta \left( \frac{M}{d} \right), \qquad
    B_r = \Theta \left( \min \left( \frac{M}{d}, \frac{M}{B_c} \right) \right) = \Theta \left( \min \left( \frac{M}{d}, d \right) \right).
  \end{equation*}
  We then have:
  \begin{equation*}
    T_c = \frac{N}{B_c} = \Theta \left( \frac{Nd}{M} \right).
  \end{equation*}

  As a result, the number of HBM accesses is:
  \begin{equation*}
    \Theta \left( Nd T_c \right) = \Theta \left( \frac{N^2 d^2}{M} \right).
  \end{equation*}

\end{proof}

\begin{proof}[Proof of \cref{thm:lower_bound}]
  For contradiction, suppose that there exists an algorithm that computes
  exact attention where the number for HBM access for all $M \in [d, Nd]$ is
  \begin{equation*}
    o \left( \frac{N^2d^2}{M} \right).
  \end{equation*}

  In the regime of $M = \Theta(Nd)$, this results in the number of HBM accesses:
  \begin{equation*}
    o \left( \frac{N^2 d^2}{Nd} \right) = o(Nd).
  \end{equation*}
  However, the input to attention (matrices $\vQ, \vK, \vV$) and the output $\vO$ have
  size $Nd$ and they start out being in HBM, so if the algorithm computes exact
  attention it must incur at least $\Omega(Nd)$ HBM accesses.
  This is a contradiction.
\end{proof}

\begin{proof}[Proof of \cref{thm:io_complexity_bwd}]
  The IO complexity of the attention backward is very similar to the IO
  complexity of the attention forward (\cref{thm:io_complexity}).
  Here we provide a sketch of the proof.

  We first analyze the IO complexity of standard attention backward pass.
  The inputs $\vQ, \vK, \vV, \vdO \in \mathbb{R}^{N \times d}$ reside in HBM, and
  the at the end of the algorithm the outputs $\vdQ, \vdK, \vdV \in \mathbb{R}^{N \times d}$ are
  written to HBM.

  At each step of the standard attention backward pass, one needs to load inputs
  of size $Nd$ or $N^2$ from HBM, and needs to write the outputs of size $N^2$
  or $Nd$ to HBM.
  This incurs $\Theta(Nd + N^2)$ HBM accesses.

  We now analyze the IO complexity of \sysname backward pass.

  Similar to~\cref{thm:io_complexity}, we see that each element of $\vK$ and
  $\vV$ is loaded from HBM once.
  Each element of $\vdK$ and $\vdV$ is only written to HBM once.
  We make $T_c$ passes over $\vQ, \vO, \vdO$, each pass loading all of
  $\vQ, \vO, \vdO$ to HBM.
  We also make $T_c$ passes over $\vdQ$, each pass reading/writing all of $\vdQ$
  from/to HBM.
  Therefore the number of HBM accesses is
  $\Theta \left( Nd + Nd T_c \right) = \Theta(Nd T_c)$.

  As in the proof of~\cref{thm:io_complexity}, the constraints on the block
  sizes are that:
  \begin{equation*}
    B_c = \Theta \left( \frac{M}{d} \right), \qquad
    B_r = \Theta \left( \min \left( \frac{M}{d}, d \right) \right).
  \end{equation*}
  We then have:
  \begin{equation*}
    T_c = \frac{N}{B_c} = \Theta \left( \frac{Nd}{M} \right).
  \end{equation*}

  As a result, the number of HBM accesses is:
  \begin{equation*}
    \Theta \left( Nd T_c \right) = \Theta \left( \frac{N^2 d^2}{M} \right).
  \end{equation*}

\end{proof}

\section{Extension Details}
\label{sec:extension_details}

\subsection{Block-sparse \sysname}
\label{subsec:block_sparse_details}

We describe the full block-sparse \sysname algorithm
in~\cref{alg:blocksparse_stream_attn}.
The algorithm is identical to~\cref{alg:fwd_full}, except that we skip zero blocks.
\iftoggle{arxiv}{
\begin{algorithm}[h]
} {
\begin{algorithm}[h]
}
  \caption{\small\label{alg:blocksparse_stream_attn}Block-Sparse \sysname Forward Pass}
  \begin{algorithmic}[1]
    \REQUIRE Matrices $\vQ, \vK, \vV \in \mathbb{R}^{N \times d}$ in HBM, on-chip SRAM of
    size $M$, softmax scaling constant $\tau \in \mathbb{R}$, masking function
    $\textsc{mask}$, dropout probability $p_\mathrm{drop}$, block sizes
    $B_c = \left \lceil \frac{M}{4d} \right\rceil, B_r = \min\left( \left \lceil \frac{M}{4d} \right\rceil, d\right)$, block sparsity mask $M \in \{ 0, 1 \}^{N/B_r \times N/B_c}$..
    \STATE Initialize the pseudo-random number generator state ${\cal R}$ and save to HBM.
    \STATE Initialize $\vO = (0)_{N \times d} \in \mathbb{R}^{N \times d}, \ell = (0)_N \in \mathbb{R}^{N}, m = (-\infty)_N \in \mathbb{R}^{N}$ in HBM.
    \STATE Divide $\vQ$ into $T_r = \left\lceil\frac{N}{B_r} \right\rceil$ blocks $\vQ_1, \dots, \vQ_{T_r}$ of size $B_r \times d$ each,
    and divide $\vK, \vV$ in to $T_c = \left\lceil \frac{N}{B_c} \right\rceil$ blocks $\vK_1, \dots, \vK_{T_c}$ and
    $\vV_1, \dots, \vV_{T_c}$, of size $B_c \times d$ each.
    \STATE Divide $\vO$ into $T_r$ blocks $\vO_i, \dots, \vO_{T_r}$ of size
    $B_r \times d$ each, divide $\ell$ into $T_r$ blocks $\ell_i, \dots, \ell_{T_r}$ of size
    $B_r$ each, divide $m$ into $T_r$ blocks $m_1, \dots, m_{T_r}$ of size $B_r$ each.
    \FOR{$1 \le j \le T_c$}
      \STATE Load $\vK_j, \vV_j$ from HBM to on-chip SRAM.
      \FOR{$1 \le i \le T_r$}
        \IF{$M_{ij} \neq 0$}
        \STATE Load $\vQ_i, \vO_i, \ell_i, m_i$ from HBM to on-chip SRAM.
        \STATE On chip, compute $\vS_{ij} = \tau \vQ_i \vK_j^T \in \mathbb{R}^{B_r \times B_c}$.
        \STATE On chip, compute $\vS_{ij}^{\mathrm{masked}} = \textsc{mask}(\vS_{ij})$.
        \STATE On chip, compute $\tilde{m}_{ij} = \mathrm{rowmax}(\vS_{ij}^{\mathrm{masked}}) \in \mathbb{R}^{B_r}$, $\tilde{\vP}_{ij} = \exp(\vS_{ij}^{\mathrm{masked}} - \tilde{m}_{ij}) \in \mathbb{R}^{B_r \times B_c}$ (pointwise),
        $\tilde{\ell}_{ij} = \mathrm{row sum}(\tilde{\vP}_{ij}) \in \mathbb{R}^{B_r}$.
        \STATE On chip, compute $m_i^{\mathrm{new}} = \max(m_i, \tilde{m}_{ij}) \in \mathbb{R}^{B_r}$, $\ell_i^{\mathrm{new}} = e^{m_i - m_i^{\mathrm{new}}} \ell_i + e^{\tilde{m}_{ij} - m_i^{\mathrm{new}}} \tilde{\ell}_{ij} \in \mathbb{R}^{B_r}$.
        \STATE On chip, compute $\tilde{\vP}_{ij}^{\mathrm{dropped}} = \mathrm{dropout}(\tilde{\vP}_{ij}, p_\mathrm{drop})$.
        \STATE Write $\vO_i \leftarrow \diag(\ell_i^{\mathrm{new}})^{-1}(\diag(\ell_i) e^{m_i - m_i^{\mathrm{new}}} \vO_i + e^{\tilde{m}_{ij} - m_i^{\mathrm{new}}}\tilde{\vP}_{ij}^{\mathrm{dropped}} \vV_j)$
        to HBM.
        \STATE Write $\ell_i \leftarrow \ell_i^{\mathrm{new}}$, $m_i \leftarrow m_i^{\mathrm{new}}$ to HBM.
        \ENDIF
      \ENDFOR
    \ENDFOR
    \STATE Return $\vO, \ell, m, {\cal R}$.
  \end{algorithmic}
\end{algorithm}

We prove the IO-complexity of block-sparse \sysname.
\begin{proof}[Proof of \cref{thm:io_complexity_blocksparse}]
  The proof is very similar to the proof of~\cref{thm:io_complexity}.
  For the block-sparse case, notice that we only need to load blocks
  corresponding to nonzero blocks.
  As a result, the number of HBM accesses are scaled by $s$, the
  fraction of nonzero blocks in the block-sparsity mask.
  However, for small values of $s$, we would still need to write the result
  $\vO \in \mathbb{R}^{N \times d}$.
  Therefore the number of HBM accesses is
  \begin{equation*}
    \Theta \left( Nd + \frac{N^2 d^2}{M} s \right).
  \end{equation*}

\end{proof}

\subsection{Potential Extensions}

We discuss here a few potential extensions of the IO-aware approach to speed up
deep learning training.

\textbf{Multi-GPU Attention.}
Large language models are trained on hundreds or thousands of GPUs, and one
typically splits the attention computation between 4-8 GPUs on the same
node~\citep{shoeybi2019megatron}.
This introduces another level of memory hierarchy: beside GPU SRAM and GPU HBM,
we also have the HBM of other GPUs.
For very long sequences, the different GPUs on the same node can cooperate to
compute attention by taking into account the asymmetry of different levels of
memory hierarchy.

\textbf{Sparse MLP layers.}
Typical dense MLP layers are compute-bound and not memory-bound.
To improve their efficiency, MLP layers with sparse weight matrices can be
used~\citep{dao2021pixelated}.
However, many sparse MLP layers are instead memory-bound, and their speedup is
often not proportional to the sparsity.
We believe that an IO-aware implementation can alleviate this issue and realize
the benefits of sparsity.
We are excited about future work in this direction, to reduce the computational
requirement of large models and improve their wall-block runtime.

\textbf{Kernel machine learning.}
Our approach in \sysname relies on the fact that the $N \times N$ attention matrix is
a function of a low-rank matrix $\vQ \vK^\top$ (of rank $d \ll N$).
As a result, we can repeatedly load the inputs $\vQ, \vK$ and recompute the
block of the attention matrix that we need, significantly reducing HBM access.
As similar scenario happens in kernel machine learning: each element $K_{ij}$ of the
$N \times N$ kernel matrix $\vK$ is a function of two vectors of size $d \ll N$, as it
measures the similarity between two datapoints $x_i$ and $x_j$.
The KeOps library~\citep{feydy2020fast,charlier2021kernel} is a successful example of how
reducing memory reads/writes can speed up kernel operations.
We hope that this will motivate kernel methods that focus more on reducing IOs
instead of just FLOPs.

\section{Full Experimental Results}
\label{sec:experiment_details}

\subsection{BERT}
\label{subsec:bert_details}

We train BERT-large following the training procedure and hyperparameters of the
reference MLPerf 1.1 implementation.
In particular, we use the LAMB optimizer with learning rate 3.75e-3, with batch
size 448, trained for at most 7100 steps.
The training is stopped once the validation accuracy (for masked language
modeling) reaches the target 72.0\%, and the wall-clock run-time is measured.
We train with FP16 precision using Apex AMP (with O2 optimization level).

We compare our results with the reported training speed from Nvidia that was
submitted to MLPerf 1.1 (\cref{table:bert_speed}).

We use the same train / validation data split provided by MLPerf 1.1 reference
implementation.
In particular, we evaluate on the same 10000 validation examples as the
baseline from Nvidia.

We train the model on 8$\times$A100-80GB GPUs. Each training run takes between 16
and 19 minutes, and we average the results of 10 runs.

\subsection{GPT-2}
\label{subsec:gpt_details}

We use the standard implementations of
GPT-2~\citep{radford2019language} from Huggingface \texttt{transformers} library and from Nvidia's Megatron-LM repo.
We follow the training recipe of the Megatron-LM repo.

We use an effective batch size of 512, and use gradient accumulation to fit into
available GPU memory.
We use the AdamW optimizer, with learning rate 6e-4 for GPT-2 small and 1.5e-4
for GPT-2 medium, and weight decay of 0.1.
All models are trained with the same hyperparameters for 400K steps.
We run all implementations with mixed-precision training (PyTorch AMP).

We use the Openwebtext dataset, with the GPT-2 BPE tokenizer. We randomly select
0.5\% of the dataset as the validation set, with the rest being used as training
set.
This random selection of validation set is done once, and all models are evaluated
on the same validation set.

We train the model on 8$\times$A100-40GB GPUs, and we measure the wall-clock training
time.
Training GPT-2 small takes between 2.7-9.5 days, and training GPT-2 medium takes
between 6.9-21.0 days (\cref{table:gpt_finetune}).

In~\cref{fig:gpt2_training_curve}, we plot of the validation perplexity throughout training of GPT-2 small/medium,
using either HuggingFace implementation or our \sysname implementation.
We see that \sysname behaves the same as the baseline implementation
and the validation perplexity curves of the two implementations almost lie on
top of each other.

\begin{figure}[ht]
  \centering
  \includegraphics[width=0.7\textwidth]{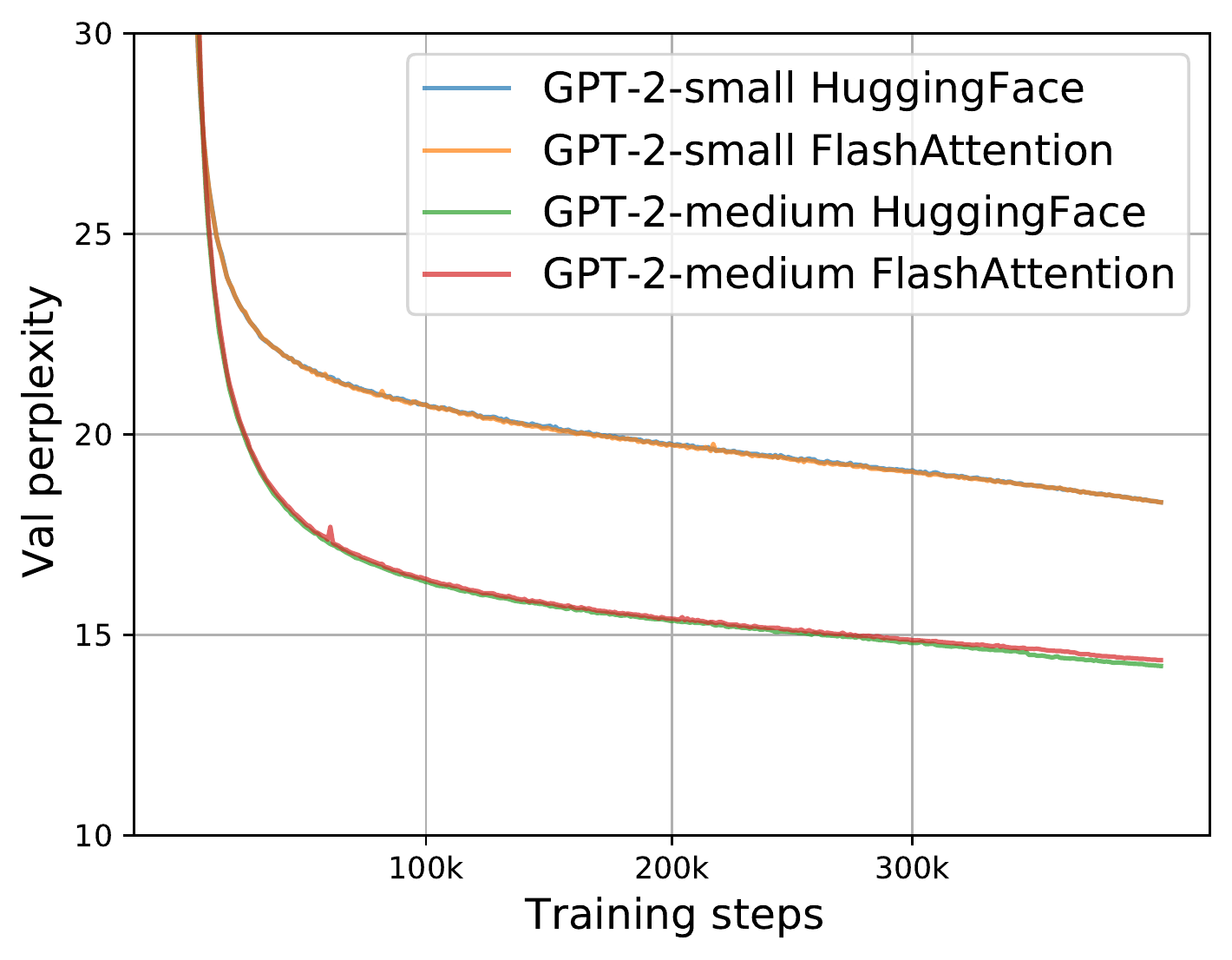}
  \caption{\label{fig:gpt2_training_curve}Validation perplexity of GPT-2
    small/medium using two implementations.
    We confirm that \sysname yields the same validation curves as the baseline
    implementation from HuggingFace.}
\end{figure}

\iftoggle{icmlworkshop}{
\subsection{Faster Transformer with \sysname on Long-range Arena}
We compare vanilla Transformer (with either standard implementation or \sysname)
on the long-range arena (LRA~\citep{tay2020long}) benchmark.
We measure accuracy, throughput, and training time of all models.
Each task has a different sequence length varying between 1024 and 4096.
We follow the implementation and experimental setting
in~\citet{tay2020long}and~\citet{xiong2021nystromformer}.\footnote{LRA accuracy
  results are known to be highly dependent on the tuning
  procedure~\citep{xiong2021nystromformer}.
  Our reproduced baselines perform better than as reported in the original
  comparison~\citep{tay2020long}.}
\cref{table:lra} shows that \sysname achieves up 2.4$\times$
speed-up compared to standard attention.
Block-sparse \sysname is faster than all of the approximate attention methods that we have
tested.

\begin{table}[h]
\captionsetup{font=small}
  \vspace{-1em}
    \caption{The performance of standard attention, \sysname, block-sparse
      \sysname, and approximate attention baselines on the Long-Range-Arena benchmarks.}
	\centering
	\small
  \iftoggle{arxiv}{}{
    \resizebox{0.9\linewidth}{!}
  }
  {
	\begin{tabular}{c|ccccc|c|c}
  Models & ListOps & Text & Retrieval & Image & Pathfinder & Avg & Speedup \\
	\hline
	Transformer & 36.0 & 63.6 & 81.6 & 42.3 & 72.7 & 59.3 & - \\
  \sysname & 37.6 & 63.9 & 81.4 & 43.5 & 72.7 & 59.8 & 2.4$\times$ \\
  Block-sparse \sysname & 37.0 & 63.0 & 81.3 & 43.6 & 73.3 & 59.6 & \textbf{2.8$\times$} \\
	\cline{1-8}
	\hline
  Linformer~\citep{wang2020linformer} & 35.6 & 55.9 & 77.7 & 37.8 & 67.6 & 54.9 & 2.5$\times$ \\
  Linear Attention~\citep{katharopoulos2020transformers} & 38.8 & 63.2 & 80.7 & 42.6 & 72.5 & 59.6 & 2.3$\times$ \\
  Performer~\citep{choromanski2020rethinking} & 36.8 & 63.6 & 82.2 & 42.1 & 69.9 & 58.9 & 1.8$\times$ \\
  Local Attention~\citep{tay2020long} & 36.1 & 60.2 & 76.7 & 40.6 & 66.6 & 56.0 & 1.7$\times$ \\
  Reformer~\citep{kitaev2020reformer} & 36.5 & 63.8 & 78.5 & 39.6 & 69.4 & 57.6 & 1.3$\times$  \\
  Smyrf~\citep{daras2020smyrf} & 36.1 & 64.1 & 79.0 & 39.6 & 70.5 & 57.9 & 1.7$\times$ \\
	\end{tabular}
  }
	\label{table:lra}
	\vspace{-1em}
\end{table}

\subsection{Better Models with Longer Sequences}
\label{ssec:exp_long_sequences}

\paragraph{Language Modeling with Long Context.}
The runtime and memory-efficiency of \sysname allow us to increase the context length of
GPT-2 by 4$\times$ while still running faster than the optimized
implementation from Megatron-LM.
\cref{table:gpt2_long_context} shows that that GPT-2 with \sysname and
context length 4K is still 30\% faster than GPT-2 from Megatron with context
length 1K, while achieving 0.7 better perplexity.

\begin{table}[h]
\vspace{-3mm}
  \captionsetup{font=small}
  \small
  \centering
  \caption{\label{table:gpt2_long_context}GPT-2 small with \sysname, with 4$\times$ larger context
    length compared to Megatron-LM, is still 30\% faster while achieving 0.7
    better perplexity. Training time on 8$\times$A100 GPUs is reported.}
  \setlength{\tabcolsep}{5pt}
  \vspace{1em}
  \iftoggle{arxiv}{}{
      \resizebox{0.8\linewidth}{!}
  }
  {
    \begin{tabular}{@{}c|ccc@{}}
      Model implementations & Context length &\multicolumn{1}{c}{OpenWebText (ppl)}&\multicolumn{1}{c}{Training time (speedup)} \\
    \hline
      GPT-2 small - Megatron-LM & 1k & 18.2 & 4.7 days (1.0$\times$) \\
      GPT-2 small - \sysname & 1k & 18.2 & \textbf{2.7 days (1.7$\times$)} \\
      GPT-2 small - \sysname & 2k & 17.6 & 3.0 days (1.6$\times$) \\
      GPT-2 small - \sysname & 4k & \textbf{17.5} & 3.6 days (1.3$\times$) \\
    \end{tabular}
  }
  \vspace{-3mm}
\end{table}
}{}

\paragraph{Long Document Classification.}
\iftoggle{icmlworkshop}{
Training Transformers with longer sequences with \sysname improves performance on the MIMIC-III~\citep{johnson2016mimic} and ECtHR~\citep{chalkidis-etal-2019-neural, chalkidis-et-al-2021-ecthr} datasets.
MIMIC-III contains intensive care unit patient discharge summaries, each annotated with multiple labels.
ECtHR contains legal cases from the European Court of Human Rights, each of which is mapped to articles of the Convention of Human Rights that were allegedly violaged.
Both of these datasets contain very long text documents; the average number of tokens in MIMIC is 2,395 tokens, and the longest document contains 14,562 tokens, while the average and longest numbers in ECtHR are 2,197 and 49,392, respectively.
We evaluate lift from increasing the sequence length of a pretrained RoBERTa model~\citep{liu2019roberta} (we repeat the positional embeddings, as in~\citet{beltagy2020longformer}).

Table~\ref{tab:mimic} shows that sequence length 16K outperforms length 512 by 4.3 points on MIMIC, and that  length 8K outperforms length 512 by 8.5 points on ECtHR.
The discrepancies may be due to subtle distribution shifts: MIMIC-III contains specialized medical text and thus may be more susceptible to a distribution shift in the document length, whereas ECtHR contains general language.

\vspace{-1em}
\begin{table}[h]
    \centering
    \begin{minipage}{2.5in}
    \small
\captionsetup{font=small}
\caption{Long Document performance (micro $F_1$) at different sequence lengths using \sysname.}
\resizebox{1.05\linewidth}{!}
{
\begin{tabular}{@{}r|ccccccccc@{}}
 & 512 & 1024 & 2048 & 4096 & 8192 & 16384 \\
\hline
MIMIC-III~\citep{johnson2016mimic} & 52.8 & 50.7 & 51.7 & 54.6 & 56.4 & \textbf{57.1} \\
ECtHR~\citep{chalkidis-etal-2019-neural} & 72.2 & 74.3 & 77.1 & 78.6 & \textbf{80.7} & 79.2 \\
\end{tabular}
}
\label{tab:mimic}

    \end{minipage}
    \begin{minipage}{0.20in}
    ~
    \end{minipage}
    \begin{minipage}{2.5in}
    \captionsetup{font=small}
    \caption{We report the first Transformer model that can achieve non-random performance on Path-X and Path-256.}
	\centering
	\small
  \resizebox{0.95\linewidth}{!}
  {
	\begin{tabular}{c|cc}
    {\bf Model}  & Path-X & Path-256 \\
	\hline
	Transformer& \xmark & \xmark \\
	Linformer~\citep{wang2020linformer}& \xmark & \xmark \\
	Linear Attention~\citep{katharopoulos2020transformers}& \xmark & \xmark \\
	Performer~\citep{choromanski2020rethinking}& \xmark & \xmark \\
	Local Attention~\citep{tay2020long} & \xmark & \xmark \\
	Reformer~\citep{kitaev2020reformer}& \xmark & \xmark \\
	SMYRF~\citep{daras2020smyrf}& \xmark & \xmark \\
	
    \hline
    \sysname & \textbf{61.4} & \xmark \\
    Block-sparse \sysname & 56.0 & \textbf{63.1} \\
	
	\end{tabular}
  }
	\label{table:pathx}

    \end{minipage}
\end{table}
\vspace{-1em}
}{}

For MIMIC-III and ECtHR, we follow the hyperparameters of~\citet{dai2022revisiting}.

\subsection{LRA details}
\label{subsec:lra_details}

We follow the hyperparameters from the Long-range arena
paper~\citep{tay2020long}, the Long-range arena repo
(\url{https://github.com/google-research/long-range-arena}), and the
Nystr{\"o}mformer reproduction~\citep{xiong2021nystromformer}.
To be generous to the baseline methods, if we are unable to reproduce the
performance of any baseline for any of the five tasks, we report the better
performance from~\citet{tay2020long} or~\citet{xiong2021nystromformer} for that
baseline on that task.

After hyperparameter tuning, almost all of the attention methods achieve similar
accuracy on all of the five LRA tasks.

We run all methods with mixed-precision training, except for Performer (not
stable with mixed precision) and Local Attention (implementation does not
support FP16).

To calculate the overall wallclock-time speedup, we take the geometric mean of
the wallclock-time speedup of each of the five tasks.

\paragraph{Path-X}
For Path-X and Path-256, we follow the hyperparameters from the PathFinder-32 experiments from the long-range arena paper\citep{tay2020long}.
For both, we first pretrain a model on Path-64.
We take the checkpoint after 200 epochs, upsample its positional embedding (we duplicate the positional embeddings gridwise in space), and fine-tune it on the downstream task for 200 epochs with one epoch of linear warmup, and cosine decay of the learning rate.
For Path-X, we take the best performing checkpoint (according to val accuracy), and additionally fine-tune it for 200 epochs with the same warmup and learning rate (this adds roughly 4 points of accuracy to \sysname for Path-X, but the model starts overfitting afterwards).

\iftoggle{icmlworkshop}{
\subsection{Benchmarking Attention}
\label{sec:benchmark}

\begin{figure}
\centering
\includegraphics[width=5.5in]{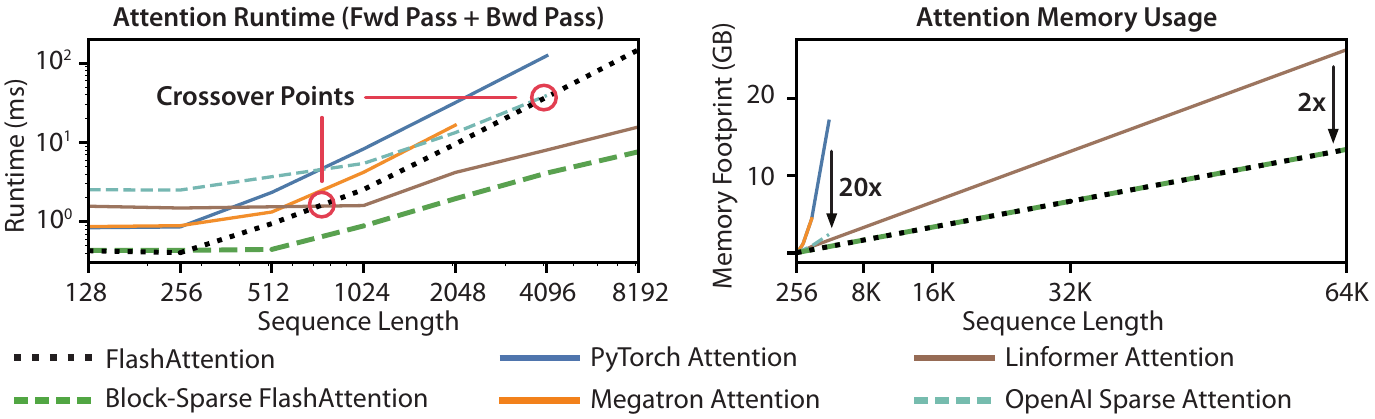}
\iftoggle{arxiv}{}{
\vspace{-1em}
}
\caption{\textbf{Left:} runtime of forward pass + backward pass. \textbf{Right:} attention memory usage.}
\label{fig:benchmark}
\iftoggle{arxiv}{}{
\vspace{-1.0em}
}
\end{figure}

We vary sequence length and measure runtime and memory usage of \sysname and block-sparse \sysname against various attention baselines on one A100 GPU with 40 GB HBM, with dropout and a padding mask.
We compare against reference implementations for exact attention, approximate attention, and sparse attention.
We report a subset of baselines in the main body; Appendix~\ref{sec:experiment_details} contains more baselines and full details.

\paragraph{Runtime.}
Figure~\ref{fig:benchmark} (left) reports the runtime in milliseconds of the forward + backward pass of \sysname and block-sparse \sysname compared to the baselines in exact, approximate, and sparse attention (exact numbers in Appendix~\ref{sec:experiment_details}).
Runtime grows quadratically with sequence length, but \sysname runs significantly faster than \textbf{exact attention} baselines, up to 3$\times$ faster than the PyTorch implementation.
The runtimes of many approximate/sparse attention mechanisms grow linearly with sequence length, but \sysname still runs faster than approximate and sparse attention for short sequences due to fewer memory accesses.
The \textbf{approximate attention} runtimes begin to cross over with \sysname at sequences between 512 and 1024.
On the other hand, block-sparse \sysname is faster than all implementations of exact, sparse, and approximate attention that we know of, across all sequence lengths.

\paragraph{Memory Footprint.}
Figure~\ref{fig:benchmark} (right) shows the memory footprint of \sysname and block-sparse \sysname compared to various exact, approximate, and sparse attention baselines.
\sysname and block-sparse \sysname have the same memory footprint, which grows linearly with sequence length.
\sysname is up to 20$\times$ more memory efficient than \textbf{exact attention} baselines, and is more memory-efficient than the \textbf{approximate attention} baselines.
All other algorithms except for Linformer run out of memory on an A100 GPU before 64K, and \sysname is still 2$\times$ more efficient than Linformer.
}{}

\subsection{Comparison with Apex FMHA}
\label{supp:fmha}

We compare our method/implementation with Apex FMHA
(\url{https://github.com/NVIDIA/apex/tree/master/apex/contrib/csrc/fmha}).

When we started this project, Apex FMHA was the fastest implementation of
attention (that we knew of), tailored for short sequences of length at most 512.
In fact, almost all MLPerf submissions for BERT training benchmark running on
Nvidia GPUs use FMHA for their model code, as of MLPerf
1.1~\citep{mattson2020mlperf}.
Since FMHA targets BERT models, it only supports head
dimension 64, and only runs on A100 GPUs.
FMHA fuses the attention computation
$\mathrm{dropout}(\softmax(\textsc{mask}(\vQ \vK^\top))) \vV$ into one CUDA kernel.
In the forward pass, it stores the attention matrix
$\softmax(\textsc{mask}(\vQ \vK^T))$ to HBM to be used in gradient computation.
As a result, it does not offer substantial memory saving (though for shorter
sequences memory footprint is often not a primary concern).

We use FMHA code as a starting point, and apply two well-established techniques
(tiling and recomputation) to deal with long sequences and to save memory as
mentioned in~\cref{sec:algo}.
As a result, we can support much longer sequences (e.g., up to length 64K).
We also support more head dimensions (16, 32, 64, 128) and broader GPU types
(all Turing and Ampere GPUs at the time of writing).

In~\cref{tab:fmha_comparison}, we compare the performance of \sysname and Apex FMHA for short sequences
(as FMHA only supports sequence length at most 512).
Generally \sysname is slightly faster than FMHA in the forward pass and slightly
slower than FMHA in the backward pass.
This is because we do not store the attention matrix in the forward pass and
recompute it in the backward pass.
Compared to FMHA, the overall runtime of \sysname is about 4\% slower for sequence length 128, 8\%
faster for sequence length 256, and 5\% faster for sequence length 512.
\begin{table}
\centering
\small
\captionsetup{font=small}
\caption{Runtime (ms) of \sysname compared to FMHA by sequence length, with
  masking and dropout, measured on an A100-SXM4-40GB GPU. Batch size 64, 16
  heads, head dimension 64 (i.e., BERT-large size).}
\begin{tabular}{@{}r|ccc@{}}
\toprule
\textbf{Attention Method} & 128 & 256 & 512 \\
\hline
\textbf{Apex FMHA forward} & 0.10 & 0.29 & 1.14 \\
\textbf{\sysname forward} & \textbf{0.08} & \textbf{0.22} & \textbf{0.81} \\
\hline
\textbf{Apex FMHA backward} & \textbf{0.17} & \textbf{0.52} & \textbf{1.81} \\
\textbf{\sysname backward} & 0.20 & 0.53 & 2.00 \\
\hline
\textbf{Apex FMHA forward + backward} & \textbf{0.27} & 0.81 & 2.95 \\
\textbf{\sysname forward + backward} & 0.28 & \textbf{0.75} & \textbf{2.81} \\
\hline
\bottomrule
\end{tabular}
\label{tab:fmha_comparison}
\end{table}

\subsection{Speedup On Different Hardware and Configurations}
\label{supp:hardware}

Speedup varies between different types of GPU types and generations depending on HBM bandwidth and SRAM size.
In this section, we profile \sysname speedup on different GPUs and configurations.

\begin{figure}[h!]
  \centering
  \includegraphics[width=5.5in]{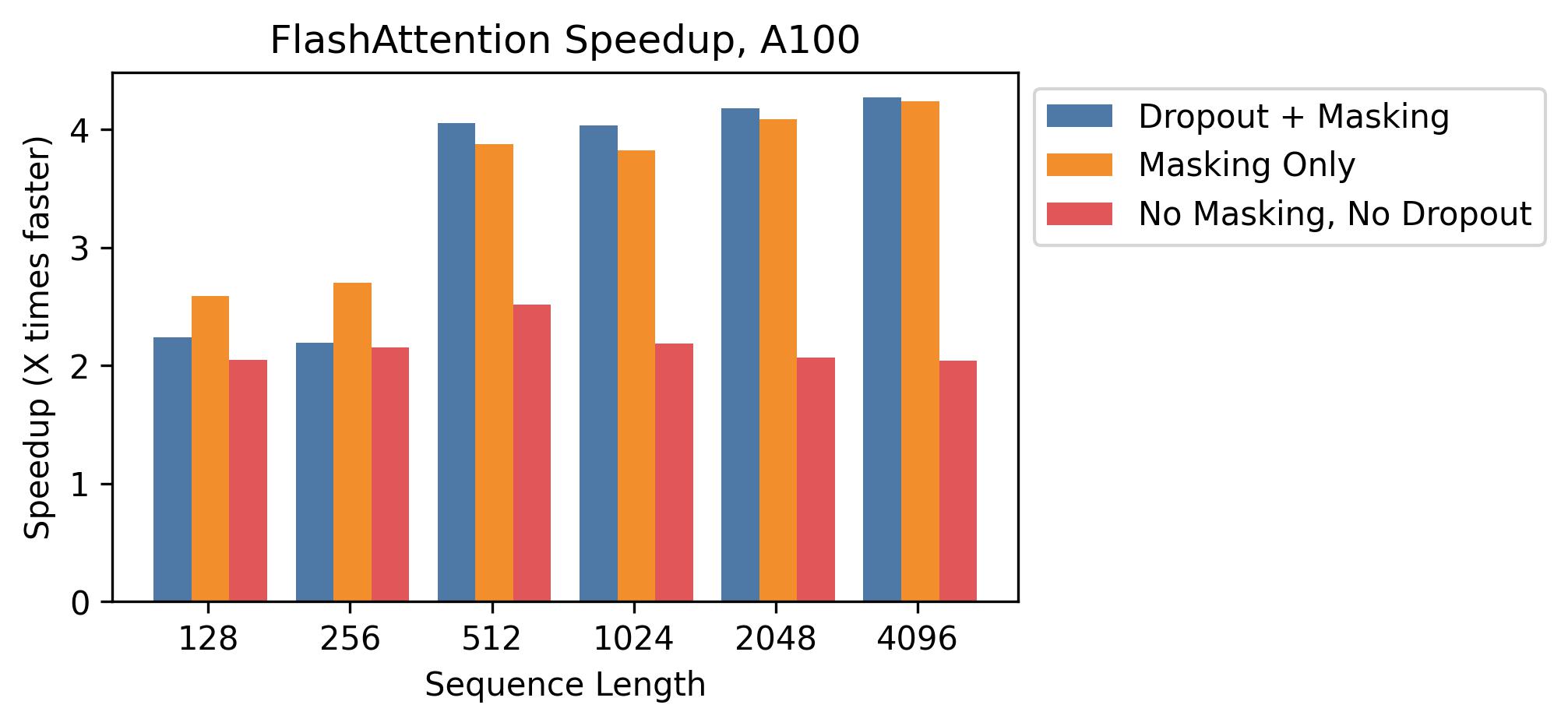}
  \caption{Speedup over standard PyTorch attention at different sequence lengths, on A100.}
  \label{fig:A100_speedup}
\end{figure}

\paragraph{A100}
Figure~\ref{fig:A100_speedup} shows speedup on an A100 GPU with batch size 8, head dimension 64, and 12 attention heads, across different sequence lengths.
We generally see 2-4$\times$ speedup, and we see more speedup when using dropout and masking due to kernel fusion.

\begin{figure}[h!]
  \centering
  \includegraphics[width=5.5in]{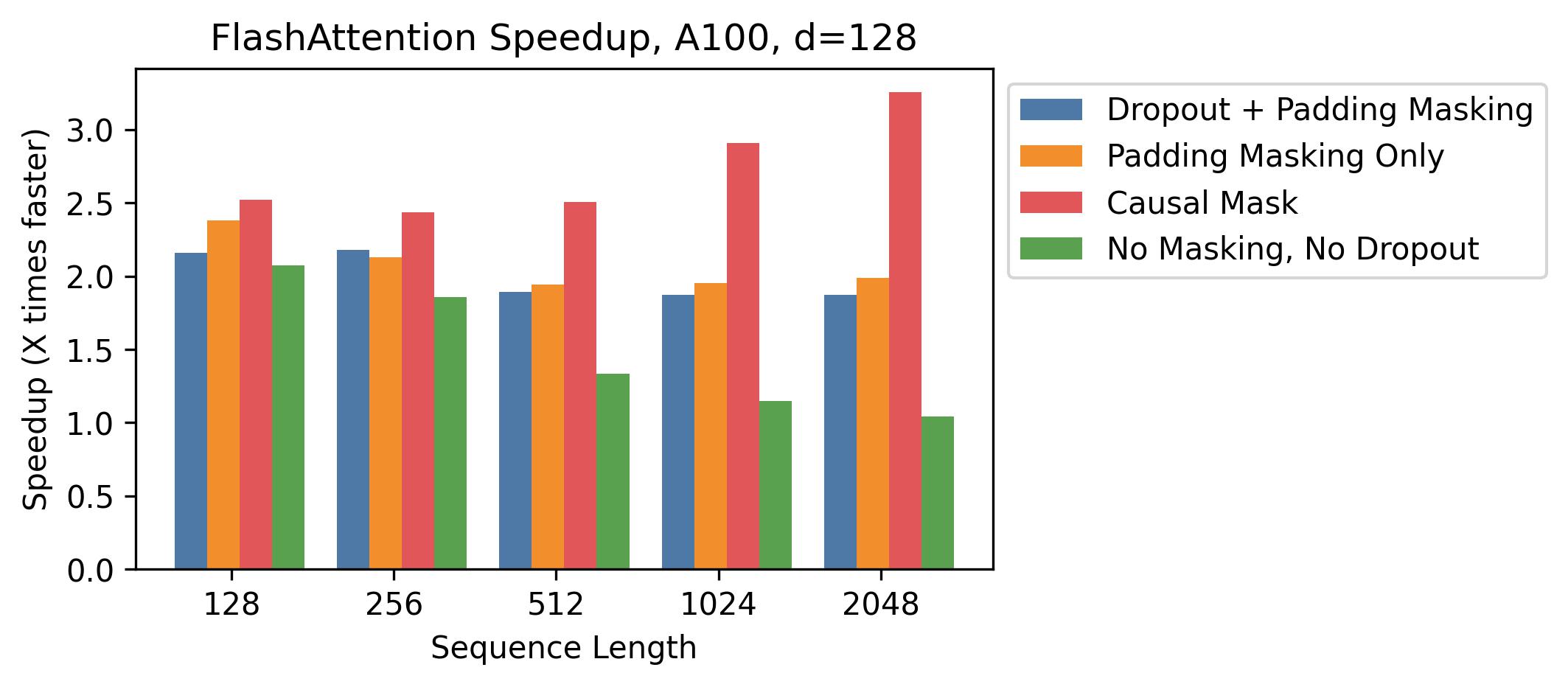}
  \caption{Speedup over standard PyTorch attention at different sequence lengths, on A100, with head dimension 128.}
  \label{fig:A100_speedup_128_dim}
\end{figure}
\paragraph{A100, Head Dimension 128}
Speedup also changes when we increase the head dimension.
Each block requires more memory, so we need to use smaller block sizes to fit into SRAM.
Figure~\ref{fig:A100_speedup_128_dim} shows speedup with head dimension 128 on an A100 (batch size 16, 12 heads).
We see less speedup overall---but we can still see significant speedup (up to 3$\times$) with a causal mask, where half the blocks are masked out.

\begin{figure}[h!]
  \centering
  \includegraphics[width=5.5in]{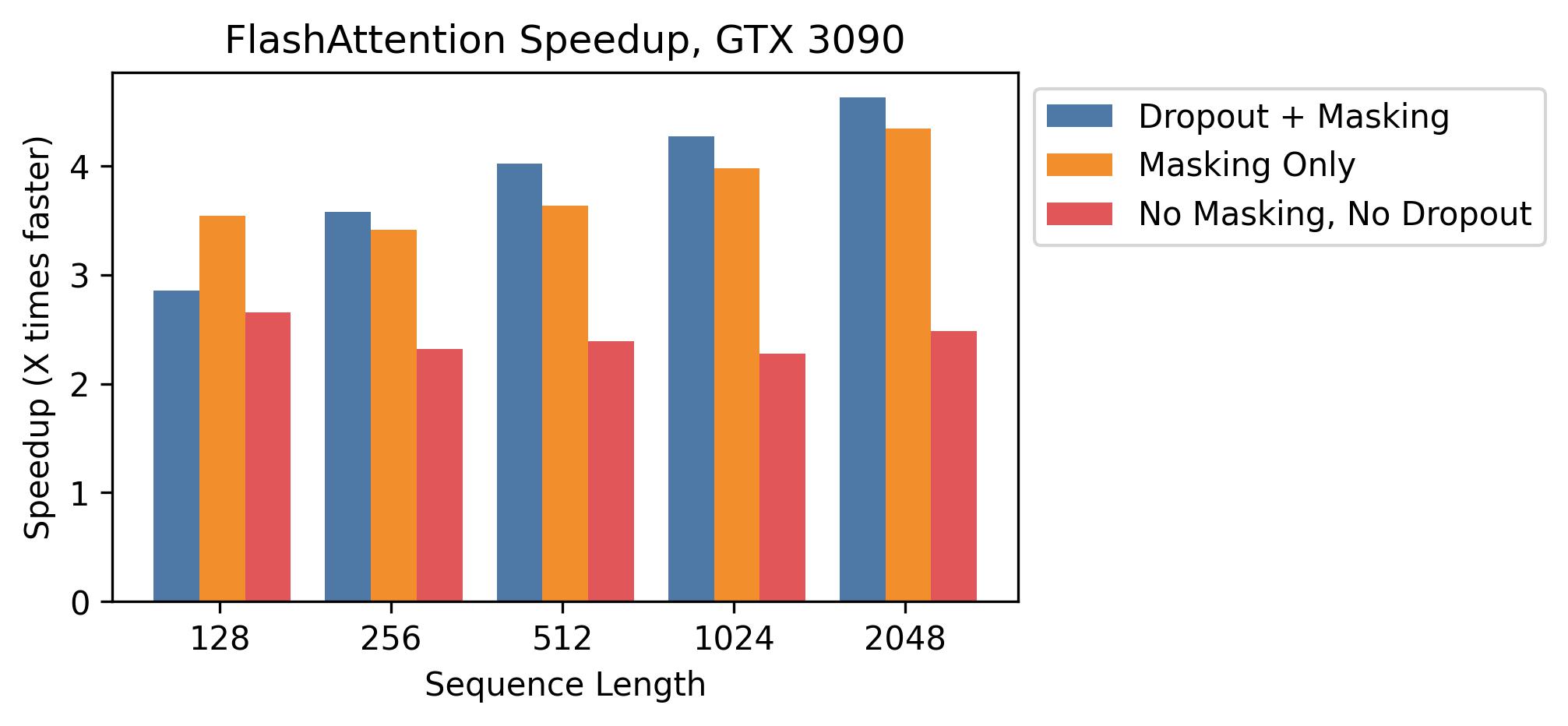}
  \caption{Speedup over standard PyTorch attention at different sequence lengths, on RTX 3090.}
  \label{fig:rtx3090_speedup}
\end{figure}

\paragraph{RTX 3090}
Figure~\ref{fig:rtx3090_speedup} shows speedup on an RTX 3090 GPU.
Here, we use batch size 12 with 12 attention heads.
We observe slightly higher speedups on the RTX 3090 (between 2.5-4.5$\times$), since the memory bandwidth on an RTX 3090 is lower than on an A100 (roughly 900 GB/s vs. 1.5 TB/s).

\begin{figure}[h!]
  \centering
  \includegraphics[width=5.5in]{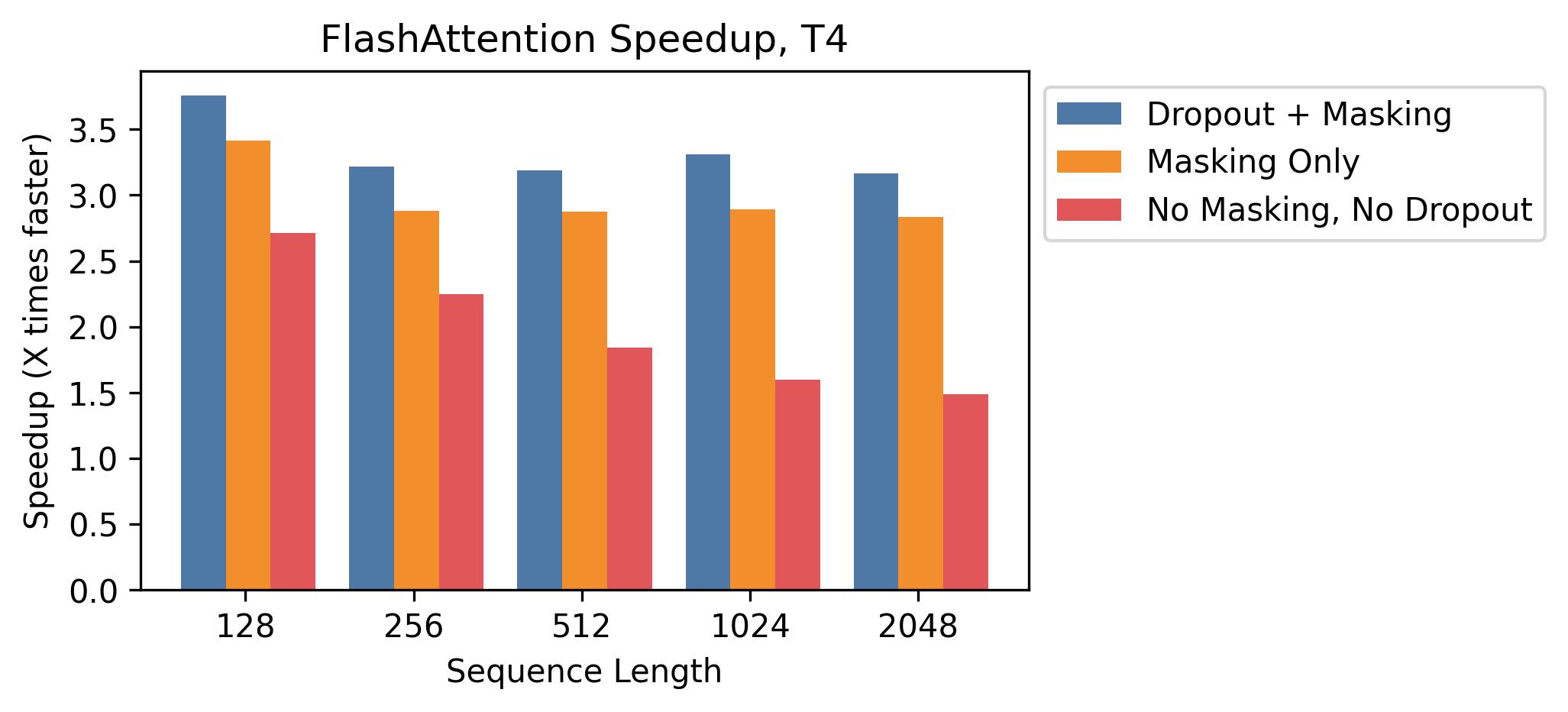}
  \includegraphics[width=5.5in]{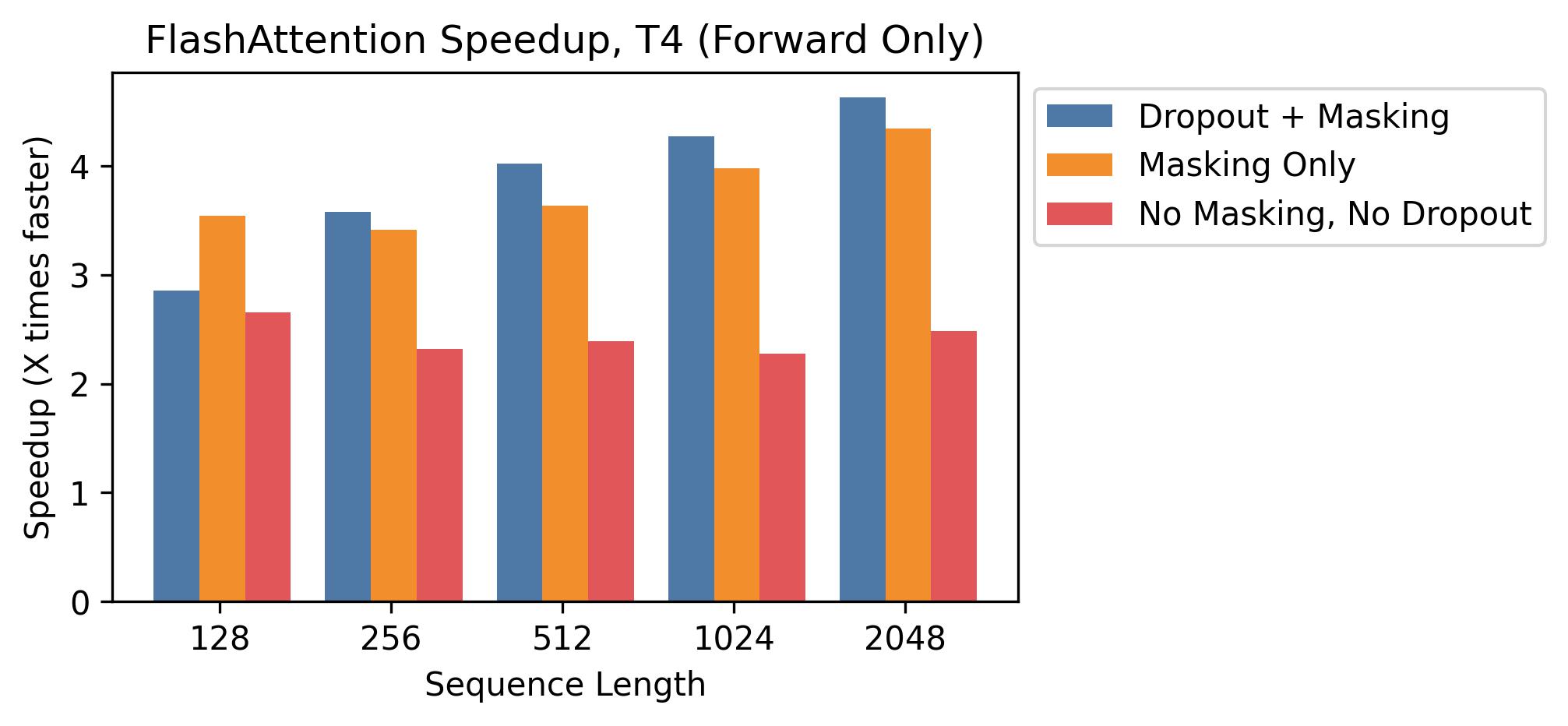}
  \caption{Speedup over standard PyTorch attention at different sequence lengths, on T4. \textbf{Top:} Combined forward pass + backward pass. \textbf{Bottom:} Forward pass only.}
  \label{fig:t4_speedup}
\end{figure}

\paragraph{T4}
Figure~\ref{fig:t4_speedup} shows speedup on a T4 GPU. T4 SRAM is smaller than A100, so we need to make the block sizes smaller in \sysname.
As a result, we observe less speedup on T4, which matches the IO complexity analysis in Section~\ref{sec:theory}.
T4 GPUs are commonly used for inference, so we also report speedup on the forward pass only.

\subsection{Full Benchmarking Results}
\label{supp:benchmarking}

We report the full benchmarking results and experimental details on A100.

\paragraph{Baselines}
We compare against reference implementations for exact attention from PyTorch/HuggingFace and Megatron, approximate attention, and sparse attention.
For approximate attention, we compare against reference implementations of Reformer~\citep{kitaev2020reformer}, Local Attention~\citep{rae-razavi-2020-transformers}, Linformer Attention~\citep{wang2020linformer}, Smyrf~\citep{daras2020smyrf}, and LongShortFormer (LSFormer)~\citep{zhu2021long}.
For sparse attention, we compare against reference implementations of Block-Sparse Attention form OpenAI~\citep{child2019generating}, Longformer\citep{beltagy2020longformer}, and BigBird Attention~\citep{zaheer2020bigbird}.
For the approximate and sparse attention, we use a compression ratio of 1/8, or a compressed sequence length of 256, whichever is smaller.

\paragraph{Setup}
We measure runtime and memory usage of the attention computation with 8 heads of dimension 64, and batch size 16 on a machine with one A100 GPU with 40 GB of GPU HBM.
We vary sequence length in our experiments.
We compute attention on random vectors for $\vQ$, $\vK$, and $\vV$ (we do not measure the projection from the hidden layer).
For dropout, we use dropout 0.1; for masking, we use a padding mask with uniformly-random mask lengths between the total sequence length and the total sequence length minus 20.
To measure runtime, we take the average of 100 measurements of the attention call.
We only measure memory footprint once, since it does not vary between runs.

We report timing results on the forward pass, backward pass, and combined forward + backward pass.
We measure each method with and without dropout, masking, or both---except for Block Sparse, Longformer, and BigBird.
These methods did not successfully run the backward pass with masking due to a bug in external libraries, so we measured them without masking to be generous.
We use FP16 for all measurements, except for Local Attention, whose
implementation only supports FP32.

For each baseline, we increase sequence length until it runs out of memory on the GPU, except for the following exceptions:
The Megatron implementation does not support sequence lengths longer than 2048.
Block-Sparse (OpenAI) does not support sequence lengths longer than 4096.
Longformer and BigBird do not support sequence lengths longer than 8092.

We measure memory usage on the combined forward + backward pass, without dropout or masking.

\paragraph{Results}
\cref{tab:benchmark_summary} summarizes all the experimental configurations and contains pointers to the results tables.

\begin{table}
    \centering
    \caption{Pointers to results tables.}
    \label{tab:benchmark_summary}
    \begin{tabular}{ccc|c}
    \toprule
        \textbf{Dropout} & \textbf{Masking} & \textbf{Pass} & \textbf{Table} \\ \hline
        Yes & Yes & Forward & \cref{tab:dropout_masking_forward_pass} \\
        Yes & Yes & Backward & \cref{tab:dropout_masking_backward_pass} \\
        Yes & Yes & Combined & \cref{tab:dropout_masking_combined} \\
        No & Yes & Forward  & \cref{tab:masking_forward_pass} \\
        No & Yes & Backward & \cref{tab:masking_backward_pass} \\
        No & Yes & Combined & \cref{tab:masking_combined} \\
        Yes & No & Forward  & \cref{tab:dropout_forward_pass} \\
        Yes & No & Backward & \cref{tab:dropout_backward_pass} \\
        Yes & No & Combined & \cref{tab:dropout_combined} \\
        No & No & Forward  & \cref{tab:forward_pass} \\
        No & No & Backward & \cref{tab:backward_pass} \\
        No & No & Combined & \cref{tab:combined} \\
        No & No & Memory Usage (Combined) & \cref{tab:memory} \\
        \toprule
    \end{tabular}
\end{table}

\begin{table}
\centering
\scriptsize
\captionsetup{font=small}
\caption{Forward pass runtime (ms) of various exact/approximate/sparse attention mechanisms by sequence length, \textbf{with dropout and masking}. Best in \textbf{bold}, second best \underline{underlined}.}
\begin{tabular}{@{}r|cccccccccc@{}}
\toprule
\textbf{Attention Method} & 128 & 256 & 512 & 1024 & 2048 & 4096 & 8192 & 16384 & 32768 & 65536 \\
\hline
\textbf{PyTorch Attention} & 0.36 & 0.34 & 0.78 & 2.54 & 9.33 & 36.33 & - & - & - & - \\
\textbf{Megatron} & 0.40 & 0.40 & 1.10 & 3.65 & 16.19 & - & - & - & - & - \\
\hline
\textbf{Reformer} & 2.03 & 3.15 & 5.67 & 11.02 & 22.59 & 46.14 & 97.38 & 212.13 & - & - \\
\textbf{Local Attention} & 0.83 & 0.86 & 1.01 & 2.20 & 7.13 & 14.32 & 28.60 & 57.79 & 117.67 & - \\
\textbf{Linformer} & 0.67 & 0.52 & 0.69 & \underline{0.71} & \underline{1.65} & \underline{3.18} & \underline{6.15} & \underline{12.16} & \underline{24.17} & \underline{52.39} \\
\textbf{Smyrf} & 2.27 & 2.34 & 3.91 & 7.44 & 14.71 & 29.22 & 58.27 & 116.41 & - & - \\
\textbf{LSformer} & 1.18 & 1.27 & 1.34 & 3.38 & 11.40 & 22.55 & 44.95 & 89.76 & 179.66 & - \\
\hline
\textbf{Block Sparse} & 1.12 & 1.11 & 2.13 & 2.77 & 6.95 & 20.91 & - & - & - & - \\
\textbf{Longformer} & 1.22 & 1.14 & 1.08 & 1.95 & 5.72 & 12.98 & - & - & - & - \\
\textbf{BigBird} & 1.13 & 1.12 & 1.12 & 1.77 & 6.03 & 13.68 & - & - & - & - \\
\hline
\textbf{\sysname} & \textbf{0.04} & \underline{0.06} & \underline{0.21} & 0.82 & 2.85 & 10.41 & 41.74 & 167.19 & 670.76 & 2682.35 \\
\textbf{Block-Sparse \sysname} & \underline{0.06} & \textbf{0.06} & \textbf{0.06} & \textbf{0.12} & \textbf{0.44} & \textbf{0.86} & \textbf{1.70} & \textbf{3.29} & \textbf{6.55} & \textbf{13.34} \\
\bottomrule
\end{tabular}
\label{tab:dropout_masking_forward_pass}
\end{table}
\begin{table}
\centering
\scriptsize
\captionsetup{font=small}
\caption{Backward pass runtime (ms) of various exact/approximate/sparse attention mechanisms by sequence length, \textbf{with dropout and masking}. Best in \textbf{bold}, second best \underline{underlined}.}
\begin{tabular}{@{}r|cccccccccc@{}}
\toprule
\textbf{Attention Method} & 128 & 256 & 512 & 1024 & 2048 & 4096 & 8192 & 16384 & 32768 & 65536 \\
\hline
\textbf{PyTorch Attention} & 0.37 & 0.49 & 1.66 & 5.81 & 22.32 & 87.67 & - & - & - & - \\
\textbf{Megatron} & 0.35 & 0.32 & 0.77 & 2.42 & 8.43 & - & - & - & - & - \\
\hline
\textbf{Reformer} & 2.37 & 4.59 & 8.91 & 17.68 & 35.13 & 70.05 & 140.01 & - & - & - \\
\textbf{Local Attention} & 0.55 & 0.62 & 1.49 & 4.03 & 13.78 & 27.61 & 55.20 & 110.27 & 221.40 & - \\
\textbf{Linformer} & 0.89 & 0.80 & 0.81 & \underline{0.93} & \underline{2.48} & \underline{4.75} & \underline{9.29} & \underline{18.27} & \underline{36.53} & - \\
\textbf{Smyrf} & 1.41 & 2.83 & 5.43 & 10.72 & 21.25 & 42.31 & 84.48 & 168.95 & - & - \\
\textbf{LSformer} & 1.75 & 1.76 & 3.01 & 7.50 & 20.07 & 39.08 & 76.39 & 150.82 & - & - \\
\hline
\textbf{Block Sparse} & 1.29 & 1.28 & 2.18 & 3.04 & 7.27 & 21.16 & - & - & - & - \\
\textbf{Longformer} & 1.27 & 1.31 & 1.29 & 2.04 & 5.24 & 10.74 & 25.95 & - & - & - \\
\textbf{BigBird} & 1.33 & 1.28 & 1.32 & 1.81 & 5.55 & 11.44 & 27.45 & - & - & - \\
\hline
\textbf{\sysname} & \textbf{0.30} & \textbf{0.26} & \underline{0.68} & 2.02 & 6.84 & 26.89 & 105.70 & 418.96 & 1666.89 & \underline{6660.44} \\
\textbf{Block-Sparse \sysname} & \textbf{0.30} & \underline{0.27} & \textbf{0.29} & \textbf{0.59} & \textbf{1.50} & \textbf{2.94} & \textbf{5.82} & \textbf{11.85} & \textbf{23.98} & \textbf{47.61} \\
\bottomrule
\end{tabular}
\label{tab:dropout_masking_backward_pass}
\end{table}
\begin{table}
\centering
\scriptsize
\captionsetup{font=small}
\caption{Forward pass + backward pass runtime (ms) of various exact/approximate/sparse attention mechanisms by sequence length, \textbf{with dropout and masking}. Best in \textbf{bold}, second best \underline{underlined}.}
\begin{tabular}{@{}r|cccccccccc@{}}
\toprule
\textbf{Attention Method} & 128 & 256 & 512 & 1024 & 2048 & 4096 & 8192 & 16384 & 32768 & 65536 \\
\hline
\textbf{PyTorch Attention} & 0.84 & 0.86 & 2.35 & 8.29 & 31.75 & 124.19 & - & - & - & - \\
\textbf{Megatron} & 0.87 & 0.89 & 1.33 & 4.21 & 16.50 & - & - & - & - & - \\
\hline
\textbf{Reformer} & 4.30 & 7.76 & 14.60 & 28.74 & 57.79 & 116.34 & 237.57 & - & - & - \\
\textbf{Local Attention} & 1.40 & 1.60 & 2.06 & 6.06 & 20.94 & 42.01 & 84.08 & 168.48 & 339.45 & - \\
\textbf{Linformer} & 1.57 & 1.49 & 1.55 & \underline{1.60} & \underline{4.19} & \underline{8.04} & \underline{15.71} & \underline{30.92} & \underline{61.47} & - \\
\textbf{Smyrf} & 3.41 & 5.08 & 9.35 & 18.18 & 36.03 & 71.68 & 143.04 & 285.87 & - & - \\
\textbf{LSformer} & 3.08 & 3.10 & 4.26 & 10.90 & 31.59 & 61.72 & 121.51 & 241.18 & - & - \\
\hline
\textbf{Block Sparse} & 2.54 & 2.52 & 3.71 & 5.44 & 13.29 & 39.19 & - & - & - & - \\
\textbf{Longformer} & 2.47 & 2.49 & 2.51 & 3.10 & 10.39 & 22.49 & 60.44 & - & - & - \\
\textbf{BigBird} & 2.51 & 2.49 & 2.52 & 3.40 & 10.97 & 23.89 & 63.28 & - & - & - \\
\hline
\textbf{\sysname} & \textbf{0.43} & \textbf{0.41} & \underline{0.95} & 2.55 & 9.56 & 37.49 & 147.75 & 586.61 & 2339.11 & \underline{9341.30} \\
\textbf{Block-Sparse \sysname} & \underline{0.44} & \underline{0.44} & \textbf{0.45} & \textbf{0.89} & \textbf{1.95} & \textbf{4.12} & \textbf{7.64} & \textbf{16.60} & \textbf{32.73} & \textbf{64.11} \\
\bottomrule
\end{tabular}
\label{tab:dropout_masking_combined}
\end{table}

\begin{table}
\centering
\scriptsize
\captionsetup{font=small}
\caption{Forward pass runtime (ms) of various exact/approximate/sparse attention mechanisms by sequence length, \textbf{with masking}. Best in \textbf{bold}, second best \underline{underlined}.}
\begin{tabular}{@{}r|cccccccccc@{}}
\toprule
\textbf{Attention Method} & 128 & 256 & 512 & 1024 & 2048 & 4096 & 8192 & 16384 & 32768 & 65536 \\
\hline
\textbf{PyTorch Attention} & 0.30 & 0.30 & 0.63 & 1.93 & 7.08 & 27.45 & 112.90 & - & - & - \\
\textbf{Megatron} & 0.45 & 0.41 & 0.43 & 1.52 & 5.80 & - & - & - & - & - \\
\hline
\textbf{Reformer} & 1.87 & 3.00 & 5.37 & 10.43 & 21.40 & 43.83 & 92.80 & 203.24 & - & - \\
\textbf{Local Attention} & 0.70 & 0.81 & 1.02 & 2.09 & 6.64 & 13.34 & 26.77 & 54.02 & 110.11 & - \\
\textbf{Linformer} & 0.63 & 0.50 & 0.67 & \underline{0.65} & \underline{1.36} & \underline{2.60} & \underline{5.04} & \underline{9.92} & \underline{19.69} & \underline{43.47} \\
\textbf{Smyrf} & 2.38 & 2.32 & 3.76 & 7.16 & 14.14 & 28.09 & 55.98 & 111.73 & - & - \\
\textbf{LSformer} & 1.22 & 1.29 & 1.44 & 3.28 & 10.99 & 21.72 & 43.29 & 86.32 & 172.76 & - \\
\hline
\textbf{Block Sparse} & 0.96 & 1.04 & 1.66 & 2.16 & 5.41 & 16.15 & - & - & - & - \\
\textbf{Longformer} & 0.99 & 0.98 & 0.99 & 1.56 & 4.79 & 11.07 & 32.98 & - & - & - \\
\textbf{BigBird} & 0.96 & 1.02 & 1.02 & 1.48 & 5.05 & 11.59 & 34.16 & - & - & - \\
\hline
\textbf{\sysname} & \textbf{0.03} & \textbf{0.04} & \underline{0.17} & 0.68 & 2.28 & 8.40 & 33.55 & 134.14 & 537.50 & 2150.88 \\
\textbf{Block-Sparse \sysname} & \underline{0.05} & \textbf{0.04} & \textbf{0.05} & \textbf{0.11} & \textbf{0.35} & \textbf{0.68} & \textbf{1.33} & \textbf{2.54} & \textbf{5.34} & \textbf{10.73} \\
\bottomrule
\end{tabular}
\label{tab:masking_forward_pass}
\end{table}
\begin{table}
\centering
\scriptsize
\captionsetup{font=small}
\caption{Backward pass runtime (ms) of various exact/approximate/sparse attention mechanisms by sequence length, \textbf{with masking}. Best in \textbf{bold}, second best \underline{underlined}.}
\begin{tabular}{@{}r|cccccccccc@{}}
\toprule
\textbf{Attention Method} & 128 & 256 & 512 & 1024 & 2048 & 4096 & 8192 & 16384 & 32768 & 65536 \\
\hline
\textbf{PyTorch Attention} & 0.44 & 0.46 & 1.53 & 5.33 & 20.34 & 79.87 & - & - & - & - \\
\textbf{Megatron} & 0.29 & 0.31 & 0.65 & 1.95 & 6.49 & - & - & - & - & - \\
\hline
\textbf{Reformer} & 2.31 & 4.47 & 8.68 & 17.20 & 34.14 & 68.09 & 136.02 & - & - & - \\
\textbf{Local Attention} & 0.51 & 0.62 & 1.30 & 3.81 & 13.33 & 26.72 & 53.41 & 106.82 & 214.15 & - \\
\textbf{Linformer} & 0.76 & 0.81 & 0.94 & \underline{0.87} & \underline{2.24} & \underline{4.25} & \underline{8.35} & \underline{16.38} & \underline{32.67} & \underline{72.11} \\
\textbf{Smyrf} & 1.34 & 2.77 & 5.30 & 10.46 & 20.73 & 41.27 & 82.41 & 164.86 & - & - \\
\textbf{LSformer} & 1.66 & 1.61 & 3.09 & 7.42 & 19.68 & 38.35 & 74.92 & 147.86 & - & - \\
\hline
\textbf{Block Sparse} & 1.24 & 1.25 & 2.04 & 2.91 & 6.78 & 19.67 & - & - & - & - \\
\textbf{Longformer} & 1.27 & 1.23 & 1.24 & 1.85 & 4.99 & 10.21 & 24.89 & - & - & - \\
\textbf{BigBird} & 1.43 & 1.50 & 1.44 & 1.69 & 5.25 & 10.86 & 26.26 & - & - & - \\
\hline
\textbf{\sysname} & \textbf{0.21} & \textbf{0.22} & \underline{0.62} & 1.84 & 5.77 & 22.25 & 86.21 & 338.91 & 1343.91 & 5361.09 \\
\textbf{Block-Sparse \sysname} & \underline{0.22} & \underline{0.22} & \textbf{0.26} & \textbf{0.57} & \textbf{1.55} & \textbf{3.13} & \textbf{5.98} & \textbf{12.21} & \textbf{23.49} & \textbf{47.85} \\
\bottomrule
\end{tabular}
\label{tab:masking_backward_pass}
\end{table}
\begin{table}
\centering
\scriptsize
\captionsetup{font=small}
\caption{Forward pass + backward pass runtime (ms) of various exact/approximate/sparse attention mechanisms by sequence length, \textbf{with masking}. Best in \textbf{bold}, second best \underline{underlined}.}
\begin{tabular}{@{}r|cccccccccc@{}}
\toprule
\textbf{Attention Method} & 128 & 256 & 512 & 1024 & 2048 & 4096 & 8192 & 16384 & 32768 & 65536 \\
\hline
\textbf{PyTorch Attention} & 0.80 & 0.81 & 2.08 & 7.23 & 27.51 & 107.58 & - & - & - & - \\
\textbf{Megatron} & 0.81 & 0.83 & 1.09 & 3.36 & 12.39 & - & - & - & - & - \\
\hline
\textbf{Reformer} & 4.16 & 7.46 & 14.06 & 27.68 & 55.66 & 112.15 & 229.37 & - & - & - \\
\textbf{Local Attention} & 1.39 & 1.68 & 2.08 & 5.83 & 20.04 & 40.16 & 80.44 & 161.35 & 325.11 & - \\
\textbf{Linformer} & 1.51 & 1.42 & 1.56 & \underline{1.67} & \underline{3.67} & \underline{6.99} & \underline{13.63} & \underline{26.77} & \underline{53.36} & \underline{117.56} \\
\textbf{Smyrf} & 3.38 & 4.93 & 9.07 & 17.66 & 34.94 & 69.55 & 138.72 & 277.41 & - & - \\
\textbf{LSformer} & 3.08 & 3.10 & 4.26 & 10.90 & 31.59 & 61.72 & 121.51 & 241.18 & - & - \\
\hline
\textbf{Block Sparse} & 2.39 & 2.40 & 3.31 & 5.02 & 12.25 & 35.94 & - & - & - & - \\
\textbf{Longformer} & 2.36 & 2.34 & 2.38 & 2.94 & 9.83 & 21.35 & 58.12 & - & - & - \\
\textbf{BigBird} & 2.35 & 2.35 & 2.37 & 3.25 & 10.36 & 22.57 & 60.63 & - & - & - \\
\hline
\textbf{\sysname} & \textbf{0.32} & \textbf{0.30} & \underline{0.83} & 2.37 & 7.95 & 30.77 & 119.98 & 473.65 & 1883.43 & 7513.01 \\
\textbf{Block-Sparse \sysname} & \underline{0.34} & \underline{0.34} & \textbf{0.36} & \textbf{0.69} & \textbf{1.85} & \textbf{3.89} & \textbf{7.16} & \textbf{14.85} & \textbf{30.46} & \textbf{60.03} \\
\bottomrule
\end{tabular}
\label{tab:masking_combined}
\end{table}

\begin{table}
\centering
\scriptsize
\captionsetup{font=small}
\caption{Forward pass runtime (ms) of various exact/approximate/sparse attention mechanisms by sequence length, \textbf{with dropout}. Best in \textbf{bold}, second best \underline{underlined}.}
\begin{tabular}{@{}r|cccccccccc@{}}
\toprule
\textbf{Attention Method} & 128 & 256 & 512 & 1024 & 2048 & 4096 & 8192 & 16384 & 32768 & 65536 \\
\hline
\textbf{PyTorch Attention} & \underline{0.26} & \underline{0.24} & 0.57 & 1.80 & 6.56 & 25.34 & - & - & - & - \\
\textbf{Megatron} & 0.27 & 0.27 & 0.56 & 1.88 & 6.56 & - & - & - & - & - \\
\hline
\textbf{Reformer} & 1.83 & 2.96 & 5.31 & 10.33 & 21.19 & 43.42 & 91.96 & 201.34 & - & - \\
\textbf{Local Attention} & 0.51 & 0.60 & 0.78 & 2.01 & 6.23 & 12.52 & 25.07 & 50.50 & 102.18 & - \\
\textbf{Linformer} & 0.47 & 0.37 & \underline{0.49} & \textbf{0.52} & \underline{1.37} & \underline{2.65} & \underline{5.12} & \underline{10.13} & \underline{20.25} & \underline{44.16} \\
\textbf{Smyrf} & 2.12 & 2.01 & 3.15 & 5.97 & 11.83 & 23.36 & 46.48 & 92.72 & - & - \\
\textbf{LSformer} & 1.28 & 1.33 & 1.51 & 3.39 & 11.40 & 22.54 & 44.96 & 89.85 & 179.73 & - \\
\hline
\textbf{Block Sparse} & 1.03 & 1.00 & 1.72 & 2.39 & 5.96 & 17.88 & - & - & - & - \\
\textbf{Longformer} & 1.02 & 1.03 & 1.03 & 1.73 & 5.10 & 11.63 & 34.22 & - & - & - \\
\textbf{BigBird} & 0.99 & 1.03 & 1.01 & 1.58 & 5.36 & 12.27 & 35.56 & - & - & - \\
\hline
\textbf{\sysname} & \textbf{0.10} & \textbf{0.10} & \textbf{0.22} & 0.83 & 2.81 & 10.38 & 41.63 & 167.01 & 668.74 & 2678.11 \\
\textbf{Block-Sparse \sysname} & 0.54 & 0.51 & 0.68 & \underline{0.61} & \textbf{0.67} & \textbf{1.10} & \textbf{1.89} & \textbf{3.71} & \textbf{7.18} & \textbf{14.41} \\
\bottomrule
\end{tabular}
\label{tab:dropout_forward_pass}
\end{table}
\begin{table}
\centering
\scriptsize
\captionsetup{font=small}
\caption{Backward pass runtime (ms) of various exact/approximate/sparse attention mechanisms by sequence length, \textbf{with dropout}. Best in \textbf{bold}, second best \underline{underlined}.}
\begin{tabular}{@{}r|cccccccccc@{}}
\toprule
\textbf{Attention Method} & 128 & 256 & 512 & 1024 & 2048 & 4096 & 8192 & 16384 & 32768 & 65536 \\
\hline
\textbf{PyTorch Attention} & 0.44 & 0.35 & 0.90 & 2.94 & 10.77 & 41.67 & - & - & - & - \\
\textbf{Megatron} & 0.28 & 0.33 & 0.92 & 2.94 & 10.80 & - & - & - & - & - \\
\hline
\textbf{Reformer} & 2.24 & 4.34 & 8.39 & 16.62 & 33.02 & 65.77 & 131.52 & - & - & - \\
\textbf{Local Attention} & 0.51 & 0.58 & 1.41 & 3.71 & 12.96 & 25.98 & 51.94 & 103.72 & 207.78 & - \\
\textbf{Linformer} & 0.84 & 0.74 & 0.79 & \underline{0.85} & \underline{2.28} & \underline{4.37} & \underline{8.66} & \underline{17.02} & \underline{33.78} & - \\
\textbf{Smyrf} & 1.27 & 2.56 & 4.90 & 9.66 & 19.16 & 38.13 & 76.17 & 152.39 & - & - \\
\textbf{LSformer} & 1.67 & 1.77 & 3.03 & 7.52 & 20.10 & 39.13 & 76.35 & 150.83 & - & - \\
\hline
\textbf{Block Sparse} & 1.27 & 1.36 & 2.15 & 3.04 & 7.27 & 21.18 & - & - & - & - \\
\textbf{Longformer} & 1.28 & 1.34 & 1.38 & 1.98 & 5.24 & 10.74 & 25.95 & - & - & - \\
\textbf{BigBird} & 1.48 & 1.47 & 1.50 & 1.81 & 5.57 & 11.38 & 27.43 & - & - & - \\
\hline
\textbf{\sysname} & \textbf{0.15} & \underline{0.18} & \underline{0.58} & 1.86 & 6.50 & 26.21 & 104.27 & 416.10 & 1661.92 & \underline{6643.01} \\
\textbf{Block-Sparse \sysname} & \underline{0.17} & \textbf{0.17} & \textbf{0.17} & \textbf{0.40} & \textbf{1.10} & \textbf{2.04} & \textbf{4.43} & \textbf{9.33} & \textbf{18.28} & \textbf{37.31} \\
\bottomrule
\end{tabular}
\label{tab:dropout_backward_pass}
\end{table}
\begin{table}
\centering
\scriptsize
\captionsetup{font=small}
\caption{Forward pass + backward pass runtime (ms) of various exact/approximate/sparse attention mechanisms by sequence length, \textbf{with dropout}. Best in \textbf{bold}, second best \underline{underlined}.}
\begin{tabular}{@{}r|cccccccccc@{}}
\toprule
\textbf{Attention Method} & 128 & 256 & 512 & 1024 & 2048 & 4096 & 8192 & 16384 & 32768 & 65536 \\
\hline
\textbf{PyTorch Attention} & \underline{0.66} & \underline{0.67} & 1.43 & 4.82 & 17.47 & 67.29 & - & - & - & - \\
\textbf{Megatron} & 0.88 & 0.90 & 1.49 & 4.73 & 17.41 & - & - & - & - & - \\
\hline
\textbf{Reformer} & 4.06 & 7.28 & 13.68 & 26.98 & 54.27 & 109.39 & 223.80 & - & - & - \\
\textbf{Local Attention} & 1.09 & 1.40 & 1.99 & 5.61 & 19.23 & 38.62 & 77.30 & 154.63 & 311.12 & - \\
\textbf{Linformer} & 1.31 & 1.21 & 1.30 & \underline{1.39} & \underline{3.73} & \underline{7.15} & \underline{14.05} & \underline{27.69} & \underline{55.00} & - \\
\textbf{Smyrf} & 3.00 & 4.37 & 8.05 & 15.66 & 31.04 & 61.64 & 123.04 & 245.65 & - & - \\
\textbf{LSformer} & 3.07 & 3.17 & 4.31 & 10.89 & 31.54 & 61.78 & 121.56 & 240.94 & - & - \\
\hline
\textbf{Block Sparse} & 2.54 & 2.52 & 3.71 & 5.44 & 13.29 & 39.19 & - & - & - & - \\
\textbf{Longformer} & 2.47 & 2.49 & 2.51 & 3.10 & 10.39 & 22.49 & 60.44 & - & - & - \\
\textbf{BigBird} & 2.51 & 2.49 & 2.52 & 3.40 & 10.97 & 23.89 & 63.28 & - & - & - \\
\hline
\textbf{\sysname} & \textbf{0.35} & \textbf{0.36} & \textbf{0.80} & 2.52 & 9.16 & 36.70 & 146.13 & 583.45 & 2332.01 & \underline{9323.63} \\
\textbf{Block-Sparse \sysname} & 0.91 & 0.83 & \underline{0.94} & \textbf{0.92} & \textbf{1.83} & \textbf{3.50} & \textbf{7.02} & \textbf{13.56} & \textbf{26.71} & \textbf{53.92} \\
\bottomrule
\end{tabular}
\label{tab:dropout_combined}
\end{table}

\begin{table}
\centering
\scriptsize
\captionsetup{font=small}
\caption{Forward pass runtime (ms) of various exact/approximate/sparse attention mechanisms by sequence length. Best in \textbf{bold}, second best \underline{underlined}.}
\begin{tabular}{@{}r|cccccccccc@{}}
\toprule
\textbf{Attention Method} & 128 & 256 & 512 & 1024 & 2048 & 4096 & 8192 & 16384 & 32768 & 65536 \\
\hline
\textbf{PyTorch Attention} & \underline{0.21} & \underline{0.22} & 0.43 & 1.27 & 4.32 & 16.47 & 67.77 & - & - & - \\
\textbf{Megatron} & 0.24 & 0.26 & \underline{0.42} & 1.33 & 4.28 & - & - & - & - & - \\
\hline
\textbf{Reformer} & 1.77 & 2.82 & 5.01 & 9.74 & 20.03 & 41.11 & 87.39 & 192.40 & - & - \\
\textbf{Local Attention} & 0.48 & 0.57 & 0.80 & 1.90 & 5.76 & 11.56 & 23.13 & 46.65 & 94.74 & - \\
\textbf{Linformer} & 0.46 & 0.36 & 0.45 & \textbf{0.50} & \underline{1.09} & \underline{2.09} & \underline{4.01} & \underline{7.90} & \underline{15.70} & \underline{35.40} \\
\textbf{Smyrf} & 1.94 & 1.96 & 3.01 & 5.69 & 11.26 & 22.23 & 44.21 & 88.22 & - & - \\
\textbf{LSformer} & 1.21 & 1.34 & 1.34 & 3.31 & 11.01 & 21.71 & 43.27 & 86.32 & 172.85 & - \\
\hline
\textbf{Block Sparse} & 0.96 & 1.04 & 1.66 & 2.16 & 5.41 & 16.15 & - & - & - & - \\
\textbf{Longformer} & 0.99 & 0.98 & 0.99 & 1.56 & 4.79 & 11.07 & 32.98 & - & - & - \\
\textbf{BigBird} & 0.96 & 1.02 & 1.02 & 1.48 & 5.05 & 11.59 & 34.16 & - & - & - \\
\hline
\textbf{\sysname} & \textbf{0.08} & \textbf{0.09} & \textbf{0.18} & 0.68 & 2.40 & 8.42 & 33.54 & 134.03 & 535.95 & 2147.05 \\
\textbf{Block-Sparse \sysname} & 0.56 & 0.52 & 0.63 & \underline{0.65} & \textbf{0.61} & \textbf{0.96} & \textbf{1.69} & \textbf{3.02} & \textbf{5.69} & \textbf{11.77} \\
\bottomrule
\end{tabular}
\label{tab:forward_pass}
\end{table}
\begin{table}
\centering
\scriptsize
\captionsetup{font=small}
\caption{Backward pass runtime (ms) of various exact/approximate/sparse attention mechanisms by sequence length. Best in \textbf{bold}, second best \underline{underlined}.}
\begin{tabular}{@{}r|cccccccccc@{}}
\toprule
\textbf{Attention Method} & 128 & 256 & 512 & 1024 & 2048 & 4096 & 8192 & 16384 & 32768 & 65536 \\
\hline
\textbf{PyTorch Attention} & 0.26 & 0.29 & 0.78 & 2.44 & 8.82 & 33.87 & - & - & - & - \\
\textbf{Megatron} & 0.29 & 0.30 & 0.80 & 2.59 & 8.86 & - & - & - & - & - \\
\hline
\textbf{Reformer} & 2.18 & 4.21 & 8.14 & 16.12 & 32.02 & 63.84 & 127.60 & - & - & - \\
\textbf{Local Attention} & 0.51 & 0.64 & 1.28 & 3.60 & 12.52 & 25.08 & 50.22 & 100.23 & 200.66 & - \\
\textbf{Linformer} & 0.69 & 0.76 & 0.69 & \underline{0.80} & \underline{2.04} & \underline{3.88} & \underline{7.67} & \underline{15.04} & \underline{30.11} & \underline{63.15} \\
\textbf{Smyrf} & 1.24 & 2.49 & 4.77 & 9.42 & 18.65 & 37.12 & 74.15 & 148.35 & - & - \\
\textbf{LSformer} & 1.68 & 1.61 & 3.02 & 7.40 & 19.72 & 38.27 & 74.89 & 147.99 & - & - \\
\hline
\textbf{Block Sparse} & 1.24 & 1.25 & 2.04 & 2.91 & 6.78 & 19.67 & - & - & - & - \\
\textbf{Longformer} & 1.27 & 1.23 & 1.24 & 1.85 & 4.99 & 10.21 & 24.89 & - & - & - \\
\textbf{BigBird} & 1.43 & 1.50 & 1.44 & 1.69 & 5.25 & 10.86 & 26.26 & - & - & - \\
\hline
\textbf{\sysname} & \textbf{0.11} & \underline{0.16} & \underline{0.52} & 1.62 & 5.45 & 21.57 & 84.75 & 336.00 & 1338.56 & 5343.19 \\
\textbf{Block-Sparse \sysname} & \underline{0.11} & \textbf{0.12} & \textbf{0.16} & \textbf{0.38} & \textbf{1.20} & \textbf{2.34} & \textbf{4.69} & \textbf{9.10} & \textbf{18.74} & \textbf{37.04} \\
\bottomrule
\end{tabular}
\label{tab:backward_pass}
\end{table}
\begin{table}
\centering
\scriptsize
\captionsetup{font=small}
\caption{Forward pass + backward pass runtime (ms) of various exact/approximate/sparse attention mechanisms by sequence length. Best in \textbf{bold}, second best \underline{underlined}.}
\begin{tabular}{@{}r|cccccccccc@{}}
\toprule
\textbf{Attention Method} & 128 & 256 & 512 & 1024 & 2048 & 4096 & 8192 & 16384 & 32768 & 65536 \\
\hline
\textbf{PyTorch Attention} & \underline{0.67} & 0.70 & 1.18 & 3.67 & 13.22 & 50.44 & - & - & - & - \\
\textbf{Megatron} & 0.74 & \underline{0.65} & 1.23 & 3.80 & 13.21 & - & - & - & - & - \\
\hline
\textbf{Reformer} & 3.93 & 7.01 & 13.15 & 25.89 & 52.09 & 105.00 & 215.13 & - & - & - \\
\textbf{Local Attention} & 1.09 & 1.27 & 1.99 & 5.38 & 18.32 & 36.77 & 73.67 & 147.29 & 296.35 & - \\
\textbf{Linformer} & 1.31 & 1.25 & 1.30 & \underline{1.29} & \underline{3.20} & \underline{6.10} & \underline{11.93} & \underline{23.39} & \underline{46.72} & \underline{100.52} \\
\textbf{Smyrf} & 2.98 & 4.23 & 7.78 & 15.12 & 29.96 & 59.45 & 118.60 & 237.02 & - & - \\
\textbf{LSformer} & 3.03 & 3.05 & 4.26 & 10.70 & 30.77 & 60.15 & 118.33 & 234.94 & - & - \\
\hline
\textbf{Block Sparse} & 2.39 & 2.40 & 3.31 & 5.02 & 12.25 & 35.94 & - & - & - & - \\
\textbf{Longformer} & 2.36 & 2.34 & 2.38 & 2.94 & 9.83 & 21.35 & 58.12 & - & - & - \\
\textbf{BigBird} & 2.35 & 2.35 & 2.37 & 3.25 & 10.36 & 22.57 & 60.63 & - & - & - \\
\hline
\textbf{\sysname} & \textbf{0.31} & \textbf{0.31} & \textbf{0.73} & 2.29 & 7.64 & 30.09 & 118.50 & 470.51 & 1876.08 & 7492.85 \\
\textbf{Block-Sparse \sysname} & 0.74 & 0.77 & \underline{0.82} & \textbf{0.88} & \textbf{1.71} & \textbf{3.21} & \textbf{6.56} & \textbf{12.60} & \textbf{24.93} & \textbf{50.39} \\
\bottomrule
\end{tabular}
\label{tab:combined}
\end{table}

\begin{table}
\centering
\scriptsize
\captionsetup{font=small}
\caption{Memory usage (MB) of various exact/approximate/sparse attention mechanisms by sequence length. Best in \textbf{bold}, second best \underline{underlined}.}
\begin{tabular}{@{}r|cccccccccc@{}}
\toprule
\textbf{Attention Method} & 128 & 256 & 512 & 1024 & 2048 & 4096 & 8192 & 16384 & 32768 & 65536 \\
\hline
\textbf{PyTorch Attention} & 36 & 104 & 336 & 1184 & 4416 & 17024 & - & - & - & - \\
\textbf{Megatron} & 36 & 104 & 336 & 1184 & 4416 & - & - & - & - & - \\
\hline
\textbf{Reformer} & 377 & 754 & 1508 & 3016 & 6033 & 12067 & 24134 & - & - & - \\
\textbf{Local Attention} & 53 & 110 & 232 & 592 & 1696 & 3392 & 6784 & 13568 & 27136 & - \\
\textbf{Linformer} & 25 & 52 & 114 & 287 & 832 & 1652 & 3292 & 6572 & 13132 & 26252 \\
\textbf{Smyrf} & 217 & 434 & 868 & 1737 & 3474 & 6947 & 13894 & 27788 & - & - \\
\textbf{LSformer} & 72 & 152 & 333 & 796 & 2540 & 5068 & 10125 & 20240 & - & - \\
\hline
\textbf{Block Sparse} & 33 & 82 & 228 & 408 & 910 & 2401 & - & - & - & - \\
\textbf{Longformer} & 30 & 61 & 124 & 277 & 681 & 1370 & 2748 & - & - & - \\
\textbf{BigBird} & 33 & 66 & 131 & 294 & 708 & 1431 & 2872 & - & - & - \\
\hline
\textbf{\sysname} & \textbf{22} & \textbf{44} & \textbf{104} & \textbf{209} & \textbf{418} & \textbf{836} & \textbf{1672} & \textbf{3344} & \textbf{6688} & \textbf{13376} \\
\textbf{Block-Sparse \sysname} & \underline{22} & \underline{44} & \underline{104} & \underline{209} & \underline{418} & \underline{836} & \underline{1672} & \underline{3344} & \underline{6690} & \underline{13384} \\
\bottomrule
\end{tabular}
\label{tab:memory}
\end{table}

\appendix
\newpage

\end{document}